\documentclass[final,3p,times,a4paper,authoryear]{elsarticle}

\usepackage{graphicx}
\usepackage{amssymb}

\usepackage{lineno}

\usepackage{url}
\usepackage{makecell}
\usepackage{bm}
\usepackage{mathtools}

\usepackage{caption}
\usepackage{subcaption}
\usepackage{tabularx}

\usepackage{diagbox}

\usepackage{xcolor}

\journal{Elsevier}

\begin{document}

\begin{frontmatter}


\title{Scalable Population Synthesis with Deep Generative Modeling}



\author{Stanislav S. Borysov}
\ead{stabo@dtu.dk}
\author{Jeppe Rich}
\author{Francisco C. Pereira}
\address{Department of Technology, Management and Economics, Technical University of Denmark, DTU, 2800 Kgs. Lyngby, Denmark}

\begin{abstract}
Population synthesis is concerned with the generation of synthetic yet realistic representations of populations. It is a fundamental problem in the modeling of transport where the synthetic populations of micro-agents represent a key input to most agent-based models. In this paper, a new methodological framework for how to 'grow' pools of micro-agents is presented. The model framework adopts a deep generative modeling approach from machine learning based on a Variational Autoencoder (VAE). Compared to the previous population synthesis approaches, including Iterative Proportional Fitting (IPF), Gibbs sampling and traditional generative models such as Bayesian Networks or Hidden Markov Models, the proposed method allows fitting the full joint distribution for high dimensions. The proposed methodology is compared with a conventional Gibbs sampler and a Bayesian Network by using a large-scale Danish trip diary. It is shown that, while these two methods outperform the VAE in the low-dimensional case, they both suffer from scalability issues when the number of modeled attributes increases. It is also shown that the Gibbs sampler essentially replicates the agents from the original sample when the required conditional distributions are estimated as frequency tables. In contrast, the VAE allows addressing the problem of sampling zeros by generating agents that are virtually different from those in the original data but have similar statistical properties. The presented approach can support agent-based modeling at all levels by enabling richer synthetic populations with smaller zones and more detailed individual characteristics.
\end{abstract}

\begin{keyword}
Population synthesis \sep Generative modeling \sep Deep learning \sep Variational Autoencoder \sep Transportation modeling \sep Agent-based modeling


\end{keyword}

\end{frontmatter}



\section{Introduction}
\label{sec:intro}

Population synthesis is a common term for methods that aim at predicting populations in social and geographical spaces under varying constraints that typically reflect profiling of the future population. The area has received increasing attention in recent years due to an increased focus on agent-based modeling in the transportation area \citep{BOWMAN20011,BRADLEY20105}. The focus on micro-agents as opposed to matrices or prototypical individuals enables modelers to investigate distributional effects in the social and geographical space \citep{Dieleman2002,BHAT2005255} and to address issues such as equity, household composition, social coherency and aging in relation to transport \citep{Stead2001,bento2005} in a more detailed fashion.

A better understanding of the contribution of the current paper and the different approaches taken in the literature may be underpinned by referring to the traditional stages in population synthesis methods, where typically three main stages are considered:
\begin{enumerate}
\item The generation of a starting solution.
\item A matrix fitting stage.
\item An allocation stage where prototypical agents are translated into micro-agents and households.
\end{enumerate}
The three-stage approach has been used in many applications \citep{BECKMAN1996415,doi:10.1002/psp.351,rich2018} and typically combines micro-surveys with deterministic matrix fitting methods and micro-simulation. In contrast, more recent simulation-based approaches to population synthesis \citep{FAROOQ2013243,SUN201549,SAADI20161,SUN2018199} tend to focus more exclusively on the first stage---the generating of appropriate representative samples for a given population  without considering how such samples can be aligned with future targets. In this paper, the focus is also entirely on the first stage of the synthesis problem. However, at the offset of the paper, it is worth noting that the relevance of a population synthesis method is its ability to produce alternative future populations typically controlled through various statistical characteristics, such as marginal distributions or combinations of them. For simulation-based approaches, this can be accomplished by defining a re-sampling stage where individuals are selected from a simulated pool of individuals. The main challenge of the re-sampling stage is to generate proper sample weights that reflect future population targets. Although this problem may not be trivial for many target variables, it can be solved by combining traditional fitting methodologies, e.g. Iterative Proportional Fitting (IPF) or simulated annealing, with quota-based random sampling.

In the population synthesis literature, there has been a trend toward probabilistic methods for the generation of starting solutions. A driver of this development has been an increasing need for more detailed populations exemplified by  smaller zones and more refined social descriptions of individuals and households. This development has underlined the importance of population synthesis, but at the same time led to new challenges related to scalability and the ability to deal with sampling zeros. Challenges with respect to scalability can refer to different things. For instance, it may refer to the fact that, from a computational perspective, there is a limit for how many attributes can be considered. However, it can also refer to the commonly known ``curse of dimensionality'' problem, which may occur if the number of modeled attributes is high. The problem is caused by the fact that the amount of data to uniformly fill a hypercube grows exponentially with the dimensionality of the hypercube. This problem is not present for low dimensions because, in that case, the distribution space is typically densely populated by data. However, for higher dimensions, when the coverage is sparser, it may lead to the existence of isolated or loosely connected ``probability density islands'' which, for example, can make sampling based on the commonly applied Markov Chain Monte Carlo (MCMC) approach less efficient. The second challenge is related to sampling zeros, that is, the agents which are missing from the original sample but exist in the real population. The term ``overfitting'' used in the paper refers to the case when a model is not capable of generating such ``out-of-sample'' agents that have similar statistical properties as those in the original sample but are not their strict copies. This paper seeks to address precisely these two challenges by proposing a new scalable approach to population synthesis based on a Variational Autoencoder (VAE) \citep{kingma2013auto} known from deep generative modeling.

Generative modeling is a subfield of statistics and machine learning which focuses on estimating the joint probability distribution of data. Traditionally, generative models based on probabilistic graphical models \citep{Bishop:2006:PRM:1162264} have suffered from scalability issues, which made them applicable either to small problems or required the introduction of simplifying assumptions. In recent years, however, deep generative models \citep{Goodfellow-et-al-2016} have opened a path toward large-scale problems. These models, which combine deep neural networks and efficient inference algorithms, have proven to be effective as a means of modeling high-dimensional data such as images \citep{radford2015unsupervised} or text \citep{SeqGANtext}. This paper aims at extending this progress into the transportation modeling area by applying the method to the problem of population synthesis.

Main contributions of the paper are as follows. First, we connect the area of population synthesis with deep generative modeling and show how a dedicated Variational Autoencoder framework can be tailored to the problem of population synthesis. In doing so, we take advantage of the scalability of the framework and the smoothing properties that result from encoding the joint distribution over a compressed latent space. Secondly, by dealing with potentially hundreds of mixed continuous and categorical variables, we take population synthesis to a new stage where it is possible to mimic entire large-scale travel diaries, including the variables that measure individual socio-economic profiles and travel preferences. Finally, we compare the proposed methodology with a conventional Gibbs sampler \citep{FAROOQ2013243} and a Bayesian Network (BN) \citep{SUN201549} by using a large-scale Danish trip diary. While these two methods outperform the VAE in a low-dimensional setting (4 attributes), they both suffer from the scalability issues when the number of modeled attributes increases (21 and 47 attributes). It is also shown that the Gibbs sampler essentially replicates the agents from the original sample when the required conditional distributions are estimated as frequency tables. At the same time, the VAE framework renders solutions that are richer and more diverse.

The rest of the paper is organized as follows. In Section 2, we briefly review existing population synthesis literature and introduce the reader to deep generative modeling. In Section 3, we describe the methodology and present a formalization of the VAE framework. Section 4 presents the case study, which is followed by Section 5 where we present results and offer a discussion. Finally, in Section 6, a conclusion is provided.

\section{Literature Review}
\label{sec:review}

\subsection{Population synthesis}
\label{sec:rev:popsynth}

Population synthesis has been approached from mainly three methodological angles: i) re-weighting, ii) matrix fitting, and iii) simulation-based approaches \citep{Tanton2014}. For the re-weighting approach, the common method is to estimate expansion factors for a given survey such that the expanded survey reflects the target population. Re-weighting typically applies non-linear optimization \citep{daly1998prototypical,bar2009estimating} to estimate weights and is generally not scalable to high dimensions. Matrix fitting can be seen as a generalization of any re-weighting approach in the sense that it renders expansion factors that can be expressed by the ratio between the starting solution and the final matrix. Popular methods for matrix fitting include Iterative Proportion Fitting (IPF) as proposed by \cite{deming1940} and maximum cross-entropy \citep{Guo2007}. There is a rich literature on these methods which oversee convergence \citep{darroch1972}, preservation of odds-ratios \citep{bishop1975discrete}, model properties under convex constraints \citep{csiszar1975,Dykstra1985} and equivalence of cross-entropy and IPF \citep{McDougall1999}. Matrix fitting, as well as re-weighting, does not produce an agent-based sample but rather a sample of prototypically weighted individuals. As a result, if the population synthesis is linked to an agent-based model, a post-simulation stage is required where individuals are drawn from the weighted sample \citep{rich2018}. The third approach to population synthesis is that of simulation. It has become increasingly popular in recent years as it can circumvent some of the shortcomings of deterministic models. Firstly, simulation-based methods can provide a more systematic way of imputing or interpolating data. Even if specific agents do not exist in the original data, it may still be possible to sample these specific agents by combining agents in the original data. Secondly, simulation-based methods often have the advantage of being effective for high-dimensional problems. These better scaling properties can address the need for more detailed populations. 

In \cite{SUN201549}, a Bayesian Network, which is essentially a probabilistic graphical model, is proposed as a means to do population synthesis. As usual in the simulation-based population synthesis literature, the idea is to model a joint distribution function of the full set of variables (agents' attributes). \cite{SUN201549} argue that if the conditional structure of the data generating process is known (i.e., the directed probabilistic graph), inference can be based on the underlying probabilities. The population can then be generated accordingly by sampling from the joint distribution. However, as this information is typically not available, \cite{SUN201549} propose that the graph structure of the data is learned through a scoring approach. The model is extended by adding hierarchical structures related to the existence of a spouse or other members in the household. Although their model performs well against IPF and a Gibbs sampler when tested on a small sample, the learning of the graph structure is known to be a computationally challenging task. This problem will be discussed further in Sections~\ref{sec:method:bn} and \ref{sec:results}.

More recently, \cite{SUN2018199} proposed an alternative hierarchical mixture modeling framework to population synthesis. In this approach, hierarchies are constituted by household attributes (e.g., family composition) and individual-specific attributes (e.g., age and gender). For each of these hierarchies, latent variables are defined. First, a variable for the household is selected and subsequently a latent variable at the individual level but conditional on the first latent variable. Based on this, the joint probability distribution of all the variables of interest is represented as the product of all the marginal distributions of the variables conditioned on the hierarchical latent variables. Inference is based on an Expectation-Maximization (EM) algorithm and the model is generally able to capture marginal distributions of all the variables and the combined joint distributions of all variables by using the latent variable representation. The challenge of this approach is not only the issue of scalability to many dimensions but also robustness with respect to the selection of the latent variables. Often it may be unclear which variables should be used as well as the implications of choosing a particular set from another. A similar modeling approach based on a probabilistic graphical model within the Bayesian formalism was considered in \cite{hu2018}.

Another method to mention is Hidden Markov Models (HMMs). The essential idea is that the observed features of a certain phenomenon are associated with hidden (latent and discrete) states of that phenomenon. As the name suggests, these states are formed in a Markovian process in the sense that each hidden state is fully determined by its previous state. This is measured by probability tables, which determine the transition rates between states. Recently, the approach has been applied to population synthesis as presented in \cite{SAADI20161}. The idea is that each latent state corresponds to an attribute. To represent all attributes for a given individual, all the attributes are sampled in sequences from the HMM. As an example, age is sampled at state 1 from a prior distribution. Then, in a second stage, the level of education is sampled from a transition table $P(\mathrm{education\ level}|\mathrm{age})$ at state 2. This process carries on for all attributes and for all individuals. However, within the proposed methodology, there is no natural ordering of variables to define the chain of states and the method resembles a Bayesian Network approach with the graph defined as a degenerate tree for the latent states.

All the mentioned above approaches can have scalability issues related to both algorithmic complexity and the curse of dimensionality. This especially becomes evident when considering performance of these models for high-dimensional problems related to image, sound or text processing, where the models based on deep artificial neural networks have recently demonstrated superior performance.

Recent progress in statistical modeling and growth of computational resources have opened a path toward estimation of complex joint distributions and the subsequent sampling from them. However, when the estimation of the full joint and/or sampling from it becomes intractable, MCMC-based approaches can be used. 

\cite{BirkinClarke1988} were the first to propose a sampling approach to population synthesis. Their methodology was to draw from conditional probability distributions of the underlying population and then gradually build a pool of individuals with a similar correlation pattern as the sample population. This is known as Gibbs sampling, which is an MCMC algorithm as also considered more recently in \cite{FAROOQ2013243}. In essence, the Gibbs sampler is a reversible Markov-Chain process which generates a chain of samples that will mimic the statistical properties of the original sample arbitrarily close \citep{Casella1992}. The Gibbs sampler simplifies the problem of estimation of the joint distribution of $n$ variables to $n$ single-variable distributions conditioned on the rest $n-1$ variables. To estimate these conditionals, statistical models can be also used, for example, discrete choice models \citep{FAROOQ2013243} or regression trees \citep{reiter2005using,Caiola:2010:RFG:1747335.1747337}. However, as will be shown in this paper, in the simplest case when these conditional probabilities are estimated based on frequency tables obtained directly from the original sample, the Gibbs sampler reproduces the original sample to perfection.

Another serious problem when using MCMC sampling approaches for high-dimensional distributions is related to the risk of being trapped in a local minimum around a starting point \citep{Justel1996}. This issue is not new and has been recognized by several authors, including \cite{gelman1992} and \cite{hills1992parameterization}. \cite{smith1993bayesian} even illustrated how the Gibbs sampler for specific bimodal distributions may be trapped when the modes of the distribution represent disconnected islands. The problem is clearly amplified in high dimensions due to the curse of dimensionality. A more detailed discussion about the Gibbs sampling as well as an illustration of its inability to avoid being trapped in a local minimum is provided in Sections~\ref{sec:method:gibbs}, \ref{sec:results} and \ref{sec:aapendixA}.

\subsection{Deep generative modeling}
\label{sec:rev:gm}

Given observations of multiple random variables, $X_1, \dots, X_n$, generative modeling aims to estimate their joint probability distribution $P(X_1, \dots, X_n)$ and/or generate new samples from it. Historically, generative models \citep{Bishop:2006:PRM:1162264} have suffered from scalability issues which has limited their use to either small-scale problems or required restrictive and simplifying assumptions.
Recently, the combination of generative models and deep learning techniques has led to efficient inference algorithms based on back-propagation \citep{Goodfellow-et-al-2016}. This has made it possible to address high-dimensional data with hundreds of attributes from the perspective of generative modeling. The most popular approaches include Variational Autoencoders (VAEs) \citep{kingma2013auto}, Generative Adversarial Networks (GANs) \citep{goodfellow2014generative} and Normalizing Flows \citep{rezende2015variational}. These models have received a great deal of attention in the computer vision field where they are used to generate photo-realistic images \citep{karras2018progressive}, to address image super-resolution \citep{DBLP:journals/corr/LedigTHCATTWS16} or for compression \citep{NIPS2016_6542} tasks. Recently, they have also gained popularity for natural language processing \citep{SeqGANtext}, speech synthesis \citep{45774}, or even in chemistry \citep{doi:10.1021/acscentsci.7b00572}, astronomy \citep{doi:10.1093/mnrasl/slx008} and physics \citep{PhysRevE.96.022140}. In the transport modeling context, deep generative models were recently applied to infer travelers' mobility patterns from mobile phone data \citep{7932990} and generate synthetic travelers' mobility patterns in order to replicate statistical properties of real people's behavior \citep{lin2017deep}. A Conditional VAE was used to study travel preference dynamics using a synthetic pseudo-panel approach \citep{borysov2019introducing}. Another application example includes generation of urbanization patterns \citep{DBLP:journals/corr/abs-1801-02710}, where synthetic maps of built-up areas are generated using the GAN model.

\section{Methodology}
\label{sec:method}

\subsection{Variational Autoencoder}
\label{sec:method:vae}

In this subsection, we provide a high-level description of the theory behind the VAE. For a more detailed discussion and mathematical rigor, we recommend readers to consider the original paper by \cite{kingma2013auto} or the tutorial by \cite{vae_tutorial}.

Let us represent an agent by a vector of random variables $X=(X_1, \dots, X_n)$ whose components can be distributed either continuously or categorically. In this case, a sample of the size $N$ represents a set of observations $\mathbf{x}_{k}$ $(k=1:N)$ where $x_{ki}$ is a $k^\mathrm{th}$ observation of the $i^\mathrm{th}$ attribute. 

The VAE is an unsupervised generative model which is capable of learning the joint distribution $P(X)$. It is a latent variable model which relates the observable variables in $X$ to a multivariate latent variable $Z$. The intuition behind the VAE is that given some known $P(Z)$, e.g., multivariate Gaussian, it can be mapped using a cleverly chosen nonlinear transformation such that it approximates $P(X)$. For instance, if $Z$ follows the two-dimensional standard normal distribution, then the values of $Z/10+Z/||Z||$ will be located on a circle (see Fig.~2 in \cite{vae_tutorial}). The samples from the VAE can be generated via sampling the latent variable and mapping it to the observed space of $X$. This mapping $P_\phi(Z)$ is referred to as a ``decoder'', where $\phi$ are its parameters. 

A na\"ive way of estimating $\phi$ is to sample values of $Z$, map them by the decoder and apply log-likelihood maximization over $\phi$ based on the observed data samples. However, this approach can be highly inefficient and suffer from very slow convergence. An alternative is to introduce structure to the latent space. This approach is based on the introduction of another function $Q_\theta(X)$ with parameters $\theta$, referred to as the ``encoder'', which maps $X$ to $Z$. During the training stage (estimation of the model's parameters, $\phi$ and $\theta$), the encoder maps an input vector to the latent space, which in turn is mapped back to the observed data space by the decoder. The error is measured as the difference between the input and output vectors. This joint estimation of $P_\phi(Z)$ and $Q_\theta(X)$ helps the model to find an efficient representation of the data in the latent space.

To approximate any complex nonlinear mapping arbitrarily close, both decoder and encoder should have a highly flexible form. The most common choice is that of an Artificial Neural Network (ANN). In this paper, we use a fully connected multilayer ANN (described below). However, any neural network architecture suitable to the task at hand can be employed, for example, a convolutional neural network for data that possess a spatial structure or a recurrent neural network for sequential data. 

In our particular case, a fully connected ANN architecture is used. It consists of multiple layers of artificial neurons. The neuron is defined in a standard way, as the weighted sum of its inputs followed by a nonlinear activation function. Namely, the $l^\mathrm{th}$ layer of an ANN computes its output as $\mathbf{y}_l=f(\mathbf{W}_{l}\mathbf{y}_{l-1} + \mathbf{b}_l)$, where $\mathbf{W}_l$ and $\mathbf{b}_l$ are the weight matrix and the bias vector associated with the layer, and $f$ is a nonlinear activation function applied element-wise to its vector argument. Further, we use the $\tanh$ function for $f$, however, other choices such as the rectified linear unit, $\mathrm{ReLU}(x)=\max(0,x)$, can be explored. The collection of all weights and biases is denoted above as $\phi$ and $\theta$ for the decoder and the encoder, respectively. For numerical variables (numerical components of the vector $X$), the output layer of the decoder has linear neurons, i.e., without applying $f$. For each categorical variable $X_i$ which can take one of the $D_i$ possible values and is represented as a one-hot vector $X_i=\left(X^{(1)}_i,\dots,X^{(D_i)}_i\right)$, $f$ in the output layer of the decoder has a soft-max form, $f(X_i)^{(m)} = \exp\left(X^{(m)}_i\right)/\sum_{j=1}^{D_i}\exp\left(X^{(j)}_i\right)$, that resembles a multinomial logit formulation.

\begin{figure}[ht!]
\centering
\includegraphics[height=4.0cm]{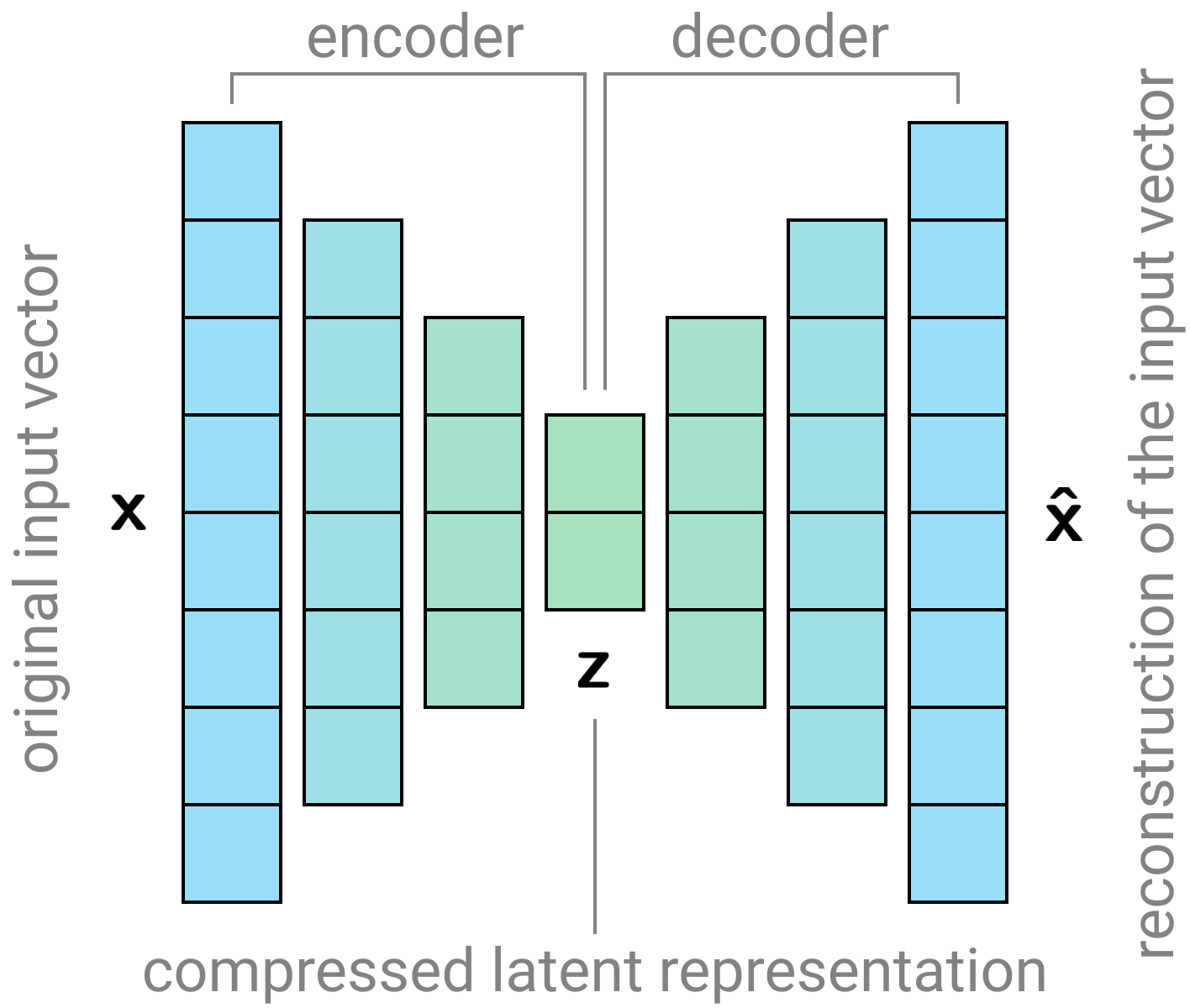}
\caption{Schematic depiction of a deterministic Autoencoder. Here, the decoder is a fully connected artificial neural network consisting of the input layer, two hidden layers and the latent representation layer. The decoder has the inverse architecture. The Autoencoder model learns to compress the input vector $\mathbf{x}$ with 8 dimensions into the latent representation $\mathbf{z}$ with 2 dimensions and reconstruct it back to the data space.}
\label{fig:ae}
\end{figure}

This encoder-decoder architecture used in the VAE is similar to the deterministic Autoencoder model \citep{Hinton504} which usually has a bottleneck structure with the dimensionality of $Z$ being much less than the dimensionality of $X$ (Fig.~\ref{fig:ae}). The encoder and the decoder usually have the same form, although such that the one is a mirror of the other. For example, if the input and the latent space dimensionality are $n$ and $D_Z$, respectively, and the encoder is an ANN consisting of the input layer, two hidden layers and the output layer with $n$-$A$-$B$-$D_Z$ neurons, then the decoder is also an ANN with the four layers consisting of $D_Z$-$B$-$A$-$n$ neurons (the ANN architecture is denoted as ``number of neurons in layer 1''-``number of neurons in layer 2''-\dots). This bottleneck structure allows for learning a compressed representation of sparse data in the low-dimensional latent space.

\begin{figure}[ht!]
\centering
\raisebox{-0.5\height}{\includegraphics[trim={0.0cm 2.0cm 0.0cm 0.0cm},clip,height=4.0cm]{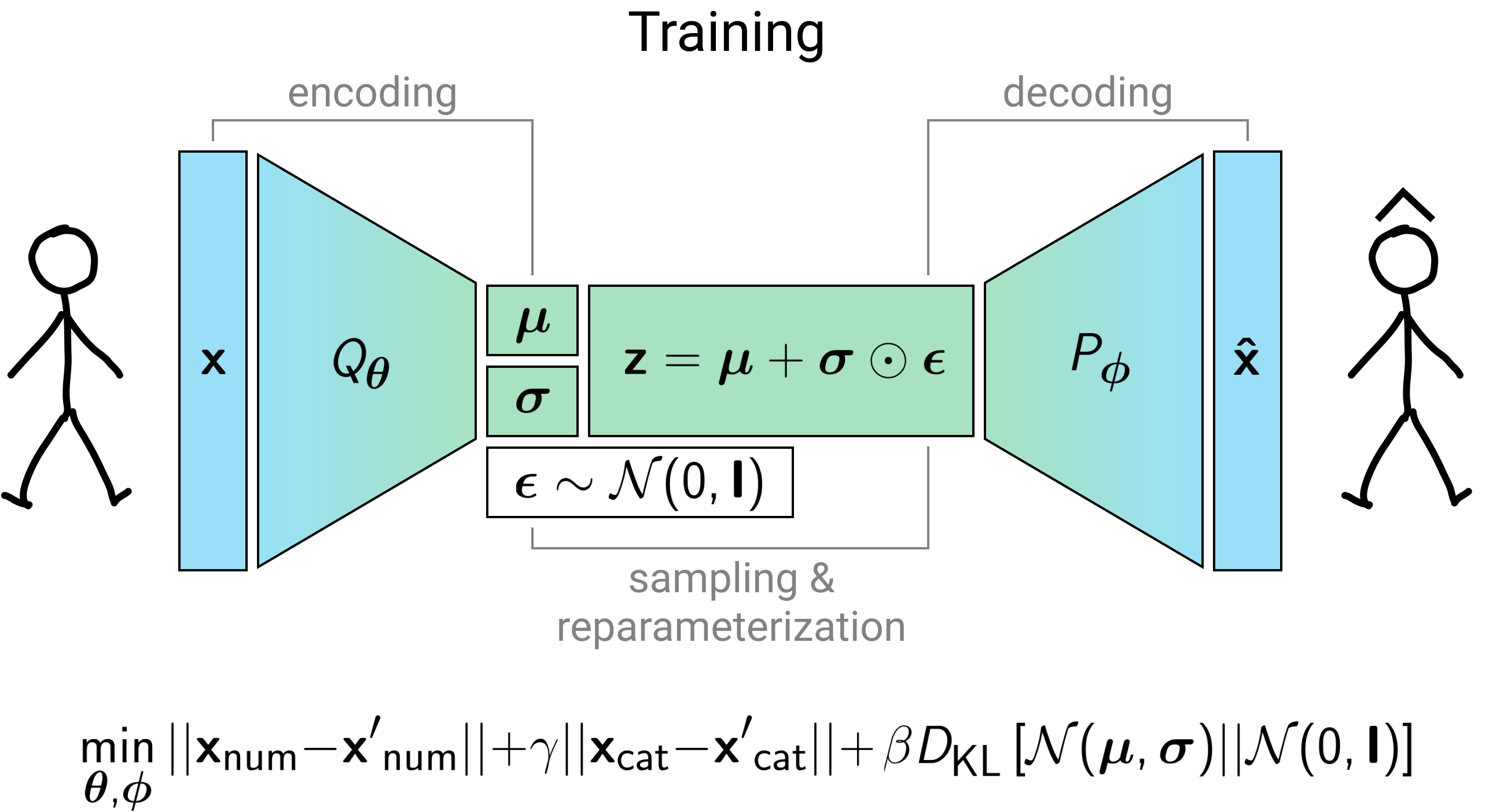}}
\hspace{0.5cm}
\raisebox{-0.47\height}{\includegraphics[trim={0.0cm 2.0cm 0.0cm 0.0cm},clip,height=3.8cm]{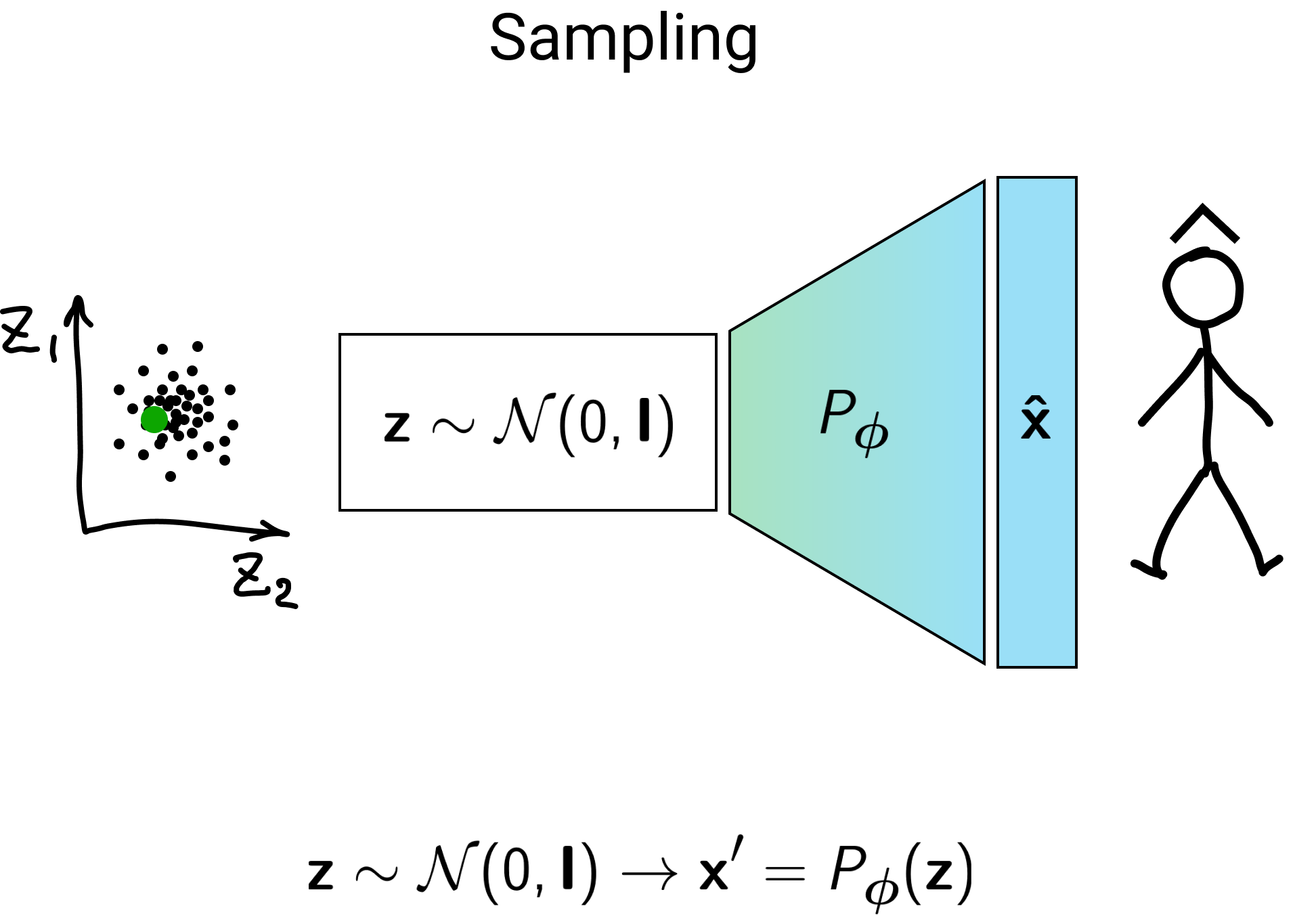}}
\caption{Schematic depiction of a Variational Autoencoder. The VAE learns to map (``decode'') the distribution of the latent variable $Z$ (for example, a multivariate Gaussian) into the data space to approximate $P(X)$. This process is facilitated by learning the latent space representation (``encoding'') of the data during the training.}
\label{fig:vae}
\end{figure}

More formally, the VAE algorithm can be summarized as follows (Fig.~\ref{fig:vae}). During the training, the encoder takes a data sample $\mathbf{x}_k$ as input and maps it into two vectors, $\bm{\mu}_k$ and $\bm{\sigma}_k$, of size $D_Z$. These vectors are used to parametrize the multivariate normal distribution for generating the latent variable $\mathbf{z}_k=\bm{\mu}_k+\bm{\sigma}_k\odot\mathbf{\epsilon}_k$, where $\mathbf{\epsilon}_k\sim\mathcal{N}(\mathbf{0},\mathbf{I}_{D_Z})$, $\mathbf{I}_{D_Z}$ is an identity matrix and $\odot$ denotes element-wise multiplication. The fact that $\mathbf{z}_k$ is not sampled directly is known as the reparameterization trick. This trick allows back-propagation of the reconstruction errors through the latent space back to the encoder. 
The back-propagation algorithm works only for continuous distributions for $Z$ although there is ongoing research that tries to overcome this limitation \citep{rolfe2016discrete,maddison2016concrete,jang2016categorical}. Then, $\mathbf{z}_k$ is transformed using the decoder, and the similarity between the original $\mathbf{x}_k$ and its reconstructed counterpart $\mathbf{\hat{x}}_k$ is measured. The following loss function is optimized using the back-propagation algorithm with respect to the encoder and decoder parameters
\begin{equation}
\label{eq:loss}
\min\limits_{\theta,\phi}\mathcal{L(\theta,\phi)} = \sum\limits_{k=1}^N||\mathbf{x}_k-\mathbf{\hat{x}}_k||_\mathrm{num} + ||\mathbf{x}_k-\mathbf{\hat{x}}_k||_\mathrm{cat} + \beta D_\mathrm{KL}\left[ \mathcal{N}(\boldsymbol{\mu}_k,\boldsymbol{\sigma}_k) || \mathcal{N}(\mathbf{0},\mathbf{I}_{D_Z}) \right],
\end{equation}
where
\begin{equation}
\label{eq:loss_num}
||\mathbf{x}_k-\mathbf{\hat{x}}_k||_\mathrm{num} = \frac{1}{2}\sum\limits_{i\in \{\mathrm{num}\}} \left(x_{ki} - \hat{x}_{ki}\right)^2
\end{equation}
is the mean square loss associated with the reconstruction of numerical variables, whereas
\begin{equation}
\label{eq:loss_cat}
||\mathbf{x}_k-\mathbf{\hat{x}}_k||_\mathrm{cat} =  - \sum\limits_{i\in \{\mathrm{cat}\}}\sum\limits_{j=1}^{D_{i}} x^{(j)}_{ki}\log \hat{x}^{(j)}_{ki}
\end{equation}
is the cross-entropy loss associated with the reconstruction of categorical variables, taking one of the $D_i$ possible values within a ``one-hot'' representation, and
\begin{equation}
\label{eq:loss_KL}
D_\mathrm{KL}\left[ \mathcal{N}(\boldsymbol{\mu}_k,\boldsymbol{\sigma}_k) || \mathcal{N}(\mathbf{0},\mathbf{I}_{D_Z}) \right] = - \frac{1}{2}\sum\limits_{i=1}^{D_Z}\left(1 + \log\sigma_{ki} - \mu_{ki}^2 - \sigma_{ki}\right)
\end{equation}
is a regularization term represented by the Kullback-Leibler divergence between the $D_Z$-dimensional latent variable distribution parametrized by the encoder output (mean $\boldsymbol{\mu}_k$ and standard deviation $\boldsymbol{\sigma}_k$) and the standard Gaussian prior $\mathcal{N}(\mathbf{0},\mathbf{I}_{D_Z})$. This term enforces the distribution of the data in the latent space to have the desired form. It makes sampling from the latent space possible in contrast to the deterministic Autoencoder model where the associated latent space lacks any probabilistic meaning. Here, $\beta$ is a hyper-parameter representing a weighting coefficient between the reconstruction losses and the regularization strength \citep{higgins2016beta}. 

When the model has been trained, new samples can be generated by sampling the latent variable from the prior distribution $\mathcal{N}(\mathbf{0},\mathbf{I}_{D_Z})$ and transforming it through the decoder to the data space. In this case, exploring new regions of the latent space can result in new out-of-sample agents that differ from those from the original data used for training. For instance, sampling values across the line connecting two points in the latent space can result in the generation of agents characterized by a mixture of attributes of the corresponding ``edge'' agents. 

The dimensions of the latent variable $Z$ can be interpretable. This behavior is more prominent for image data, where varying one of the latent dimensions, the others being fixed, might reveal that its value is correlated with the object's position, color or size. However, these properties are not so obvious in the case of non-visual data such as those considered in the present work. The VAE model can also address anomaly detection and data imputation problems. For example, the probability of an agent to be present in the data can be estimated in the latent space using the agent's mapping via the encoder. Missing values for attributes of an agent can be imputed by minimizing the distance between the known attributes of the agent and its reconstructed counterpart by using gradient descent in the latent space.

\subsection{Gibbs sampling}
\label{sec:method:gibbs}

MCMC methods are a set of popular tools to facilitate both sampling from complex distributions and performing statistical inference. They are widely used in many domains including computer science, machine learning, physics, and finance. Gibbs sampling is one of many MCMC algorithms that works by sampling from the conditional distributions $P(x_i|x_{-i})\equiv P\left(x_i|x_1,\dots,x_{i-1},x_{i+1},\dots,x_{n}\right)$. 
The process is equivalent to a sampling scheme from the joint distribution such that $P(x_i|x_{-i})=P(x_i,x_{-i})/P(x_{-i})\propto P(x_i,x_{-i})$. The standard algorithm includes two fundamental steps. First, a starting point $\mathbf{x}^{(k)}$ is defined. In the second step, the value of each component $x_i^{(k)}$ is updated according to $P\left(x_i|x_1^{(k+1)},\dots,x_{i-1}^{(k+1)},x_{i+1}^{(k)},\dots,x_{n}^{(k)}\right)$ until all the components are updated and a new sample $\mathbf{x}^{(k+1)}$ is obtained. The process is repeated using this new sample as a starting point to obtain the next sample $\mathbf{x}^{(k+2)}$. $\mathbf{x}^{(0)}$ can be initialized either randomly or from any sample available. Typically, to reach a stationary state, a ``warm-up'' stage is carried out. This stage is accomplished by eliminating the first $N_\mathrm{wu}$ samples from the sample pool. Also, to avoid correlation between samples, each consequent $N_\mathrm{th}$ sample is kept while samples in between are eliminated.

In the simplest case, the conditional probabilities required for the Gibbs sampler can be provided in the form of multidimensional frequency tables. Another possibility is to use statistical models estimated from the data, for example, discrete choice models \citep{FAROOQ2013243}, probabilistic graphical models, regression trees \citep{reiter2005using,Caiola:2010:RFG:1747335.1747337} or ANNs. In this paper, we focus on the simplest case when the frequency tables are estimated from the original disaggregated data. However, this choice does not bring any advantages over the original sample itself. The most important problem is that, in this case, the Gibbs sampler will reproduce the underlying disaggregated data due to the zero sampling probabilities for sampling zeros. The sampling procedure effectively simplifies to uniform expansion of the pool of initial agents, reproducing the underlying population to perfection. In addition to the overfitting, the Gibbs sampler can be also sensitive to the curse of dimensionality. As mentioned in Section~\ref{sec:review}, it can become trapped in a local minimum or suffer from super slow convergence. We illustrate these issues in Section~\ref{sec:results} and \ref{sec:aapendixA}.

As these problems can become worse in higher dimensions, one way to tackle them is to use ``collapsed'' Gibbs sampling \citep{doi:10.1080/01621459.1994.10476829}, where partial conditionals $P(x_i|x_{-i,-j,\dots})$ are used. The benefits of the collapsed Gibbs sampler come at the cost of neglecting statistical dependencies between the marginalized and remaining variables. However, there is no clear way to define the variables to be marginalized in the population synthesis context. In real-world applications, partial conditionals can be also used when the full conditionals are not available \citep{FAROOQ2013243}. Another way to deal with local minima is to generate several chains from different starting points. However, this approach quickly becomes infeasible in higher dimensions, where the number of starting points to cover the space grows exponentially with the number of the variables involved.

\subsection{Bayesian Network}
\label{sec:method:bn}

Here, a short explanation of the concept of a Bayesian network (BN) is provided. For more details about the application of BNs to population synthesis, the reader can refer to \cite{SUN201549}.

\begin{figure}[ht!]
\centering
(a) \hspace{3.3cm} (b)\\
\vspace{0.3cm}
\includegraphics[height=2.0cm]{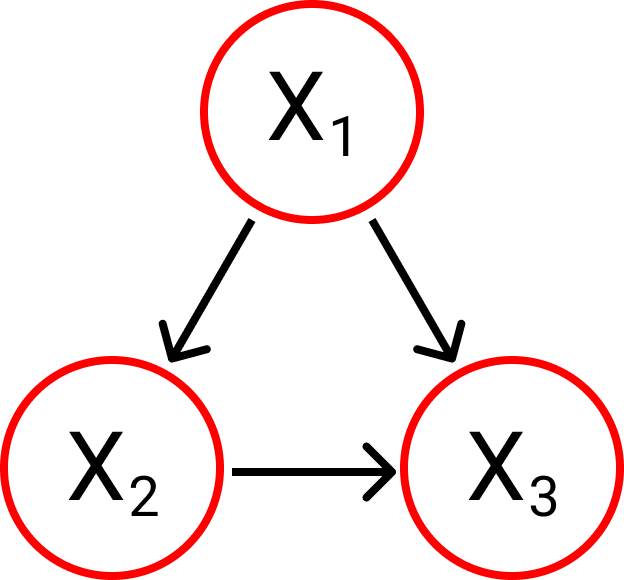}
\hspace{1.5cm}
\includegraphics[height=2.0cm]{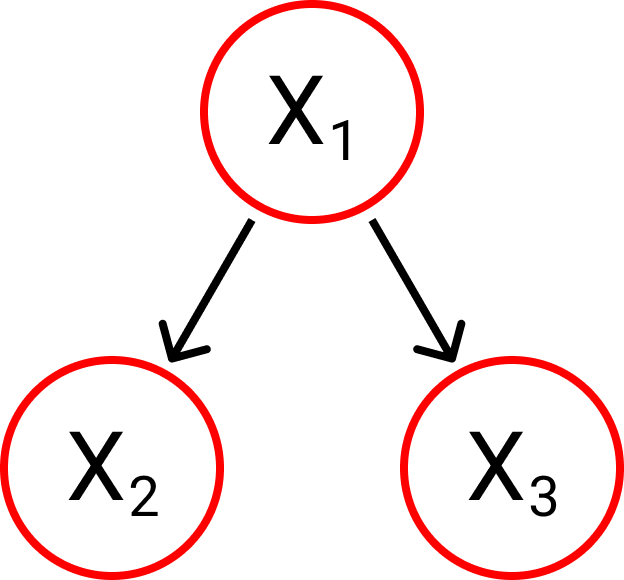}
\caption{Examples of Bayesian Networks corresponding to the two different factorizations of the joint of three variables $P(X_1,X_2,X_3)$: (a) $P(X_1)P(X_2|X_1)P(X_3|X_1,X_2)$ and (b) $P(X_1)P(X_2|X_1)P(X_3|X_1)$.}
\label{fig:bn}
\end{figure}

A Bayesian Network (BN) is a probabilistic graphical model which represents factorization of the joint probability distribution into conditional distributions represented by a directed acyclic graph (DAG). For example, the full factorization of the joint distribution of three variables $P(X_1,X_2,X_3)=P(X_1)P(X_2|X_1)P(X_3|X_1,X_2)$ corresponds to the BN depicted in Fig.~\ref{fig:bn}a. For categorical variables, these conditionals are usually represented as frequency tables, while dealing with numerical variables usually requires specification of their distribution function. However, such a factorization is not informative because it essentially overfits the original data. In order for the model to generalize, the DAG can be made sparser by dropping some conditional dependencies using a scoring function.

The scoring function represents a trade-off between the ability of the model to reproduce the original data and the complexity of the underlying DAG. For example, the widely used minimum description length (MDL) scoring, being tightly related to the Bayesian information criterion (BIC), corresponds to the data log-likelihood minus a term which is proportional to the number of free parameters in the model. For instance, such a scoring function might be in favor of dropping $P(X_3|X_1,X_2)$ dependence on $X_2$ and thereby yielding the sparser factorization $P(X_1,X_2,X_3)=P(X_1)P(X_2|X_1)P(X_3|X_1)$ as depicted in Fig.~\ref{fig:bn}b.

Learning exact BN structure using a scoring function has super-exponential complexity as it involves the evaluation of all possible DAGs. Using the dynamic programming approach helps to reduce the complexity to exponential. It can be further sped up by the A* algorithm \citep{Yuan:2011:LOB:2283696.2283763} which also yields the globally optimal graph. However, it can still be prohibitively slow when more than a few dozens of variables are considered. The complexity can be reduced to polynomial using various approximate methods including hill-climbing, greedy search and MCMC methods \citep{Heckerman1998}. For example, the greedy search includes each node iteratively to the graph when the score is improved. Also, some prior assumptions about the structure can be imposed such as restricting the number of parents / children for each node or incorporating known dependencies between the variables. For example, by requiring the DAG to have a tree structure, it can be learned with the Chow-Liu algorithm \citep{1054142} which has quadratic complexity. However, using search heuristics and assumptions usually lead to worse performance of the model.

Within the population synthesis context, a BN approach was first proposed by \cite{SUN201549} where a small-scale problem involving 7 attributes was considered. However, its scaling properties for this task have not been thoroughly investigated.

\section{Case study}
\label{sec:case}
We consider the problem of generating a synthetic population with both numerical (e.g. income) and categorical (e.g. gender or education level) attributes on the basis of a large-scale trip diary in Denmark. We use the Danish National Travel Survey (TU) data\footnote{\url{http://www.modelcenter.transport.dtu.dk/english/tvu}}, which is known to be one of the largest coherent trip diaries worldwide. It contains 24-hour trip diaries for more than 330,000 Danish residents and provides detailed information for approximately one million trips in the period from 1992 to present. In this paper, we use the TU sample from 2006--2017, representing more than 146,000 unique respondents. Given the Denmark population as of 2018, which is approximately 5.75 million people, this sample represents approximately 2.5\% of the total population, which can be regarded as a micro-sample typically available to the researchers.

As the data contain both numerical and categorical variables, we consider two approaches: i) discretization of the numerical variables by converting them into categorical variables using uniform bins which number is defined in the third column in Table~\ref{tab:data} and ii) working directly with the mixture of both. While the latter makes fewer assumptions about the variables, the former can approximate the continuous case depending on the resolution of the discretization bins. We use a ``one-hot'' encoding of the categorical variables. For example, if a discretized number of persons in the household can take one of the categorical values ``1'', ``2'', ``3'', ``4'' or ``5+'', then the household with 2 persons will correspond to the vector $(0, 1, 0, 0, 0)$. Hence, the dimensionality of the problem is directly proportional to the number of possible categories. We selected three subsets of individual attributes for each person defined in Table~\ref{tab:data}:
\begin{enumerate}
\item \emph{Basic} attributes are similar to the case study used in \cite{FAROOQ2013243} and contain 2 numerical (person's age and a number of people in the household) and 2 categorical (person's gender and educational attainment) variables. The resulting distribution is however 14-dimensional (and 27-dimensional when the numerical variables are discretized).
\item \emph{Socio} attributes contain more detailed socio-demographic information and comprise 8 numerical and 13 categorical variables, where the resulting distribution is 68-dimensional (121-dimensional when the numerical variables are discretized).
\item To investigate scalability of the methods, \emph{Extended} attributes additionally include data related to work and travel behavior and contain all 47 variables, 21 numerical and 26 categorical, making the resulting distribution 230-dimensional (357-dimensional when the numerical variables are discretized). Although such attributes are typically not present in the population synthesis problem, they can be still used in different stages of transport modeling and simulation.
\end{enumerate}

\begin{table}[ht!]
\footnotesize
\centering
\begin{tabular}{lllll}
\hline
\# & Name & Type & Number of values & Description \\
\hline
\hline 
1 & HousehNumPers & numerical (int) & 5 & Number of persons in the household \\
2 & RespAgeCorrect & numerical (int) & 8 & Age \\
3 & RespEdulevel & categorical & 12 & Educational attainment \\
4 & RespSex & binary & 2 & Gender \\
\hline
5 & Handicap & categorical & 3 & Handicap \\
6 & HousehAccomodation & categorical & 7 & Home, type \\
7 & HousehAccOwnorRent & categorical & 4 & Home, ownership \\
8 & HousehNumAdults & numerical (int) & 5 & Number of adults in the household \\
9 & HousehNumcars & numerical (int) & 5 & Number of cars in the household \\
10 & HousehNumDrivLic & numerical (int) & 5 & Number of persons with driving licence in the household \\
11 & HousehNumPers1084 & numerical (int) & 5 & Number of persons 10-84 years in the household \\
12 & IncHouseh & numerical (int) & 10 & The household's gross income \\
13 & IncRespondent & numerical (int) & 10 & The respondent’s gross income \\
14 & NuclFamType & categorical & 5 & Nuclear family type \\
15 & PosInFamily & categorical & 5 & Position in the nuclear family \\
16 & RespHasBicycle & categorical & 3 & Bicycle ownership \\
17 & ResphasDrivlic & categorical & 5 & Driver licence \\
18 & RespHasRejsekort & categorical & 3 & Danish PT smartcard \\
19 & RespHasSeasonticket & categorical & 2 & Season ticket (public transport) \\
20 & RespIsmemCarshare & categorical & 3 & Member of car sharing scheme \\
21 & RespMainOccup & categorical & 14 & Principal occupation \\
\hline
22 & DayJourneyType & categorical & 7 & Journey type of the day \\
23 & DayNumJourneys & numerical (int) & 5 & Number of journeys during 24 hours \\
24 & DayPrimTargetPurp & categorical & 28 & Primary stay of the day, purpose \\
25 & DayStartJourneyRole & categorical & 3 & Start of the day: position in journey \\
26 & DayStartPurp & categorical & 26 & Purpose at start of the day \\
27 & GISdistHW & numerical (cont) & 10 & “Bird flight” distance between home and place of occupation \\
28 & HomeAdrDistNearestStation & numerical (cont) & 4 & Home, distance to nearest station \\
29 & HomeParkPoss & categorical & 20 & Parking conditions at home \\
30 & HwDayspW & numerical (int) & 7 & Number of commuter days \\
31 & HwDaysReason & categorical & 8 & Reason for fewer commuter days \\
32 & JstartDistNearestStation & numerical (cont) & 4 & Journey base, distance to nearest station \\
33 & JstartType & categorical & 5 & Journey base, type \\
34 & ModeChainTypeDay & categorical & 14 & Transport mode chain for the entire day \\
35 & NumTripsCorr & numerical (int) & 4 & Number of trips \\
36 & PrimModeDay & categorical & 23 & Primary mode of transport for the entire day \\
37 & RespNotripReason & categorical & 7 & Reason for no trips \\
38 & TotalBicLen & numerical (cont) & 6 & Total bicycle travel distance \\
39 & TotalLen & numerical (cont) & 5 & Total travel distance of trips \\
40 & TotalMin & numerical (cont) & 5 & Total duration of trips \\
41 & TotalMotorLen & numerical (cont) & 5 & Total motorised travel distance \\
42 & TotalMotorMin & numerical (cont) & 5 & Total motorised duration of trips \\
43 & WorkatHomeDayspM & numerical (int) & 6 & Days working from home \\
44 & WorkHoursPw & numerical (int) & 8 & Number of weekly working hours \\
45 & WorkHourType & categorical & 5 & Planning of working hours \\
46 & WorkParkPoss & categorical & 12 & Parking conditions at place of occupation \\
47 & WorkPubPriv & categorical & 4 & Public- or private-sector employee \\
\hline
\end{tabular}
\caption{Individual attributes of the TU participants used in the paper. For the numerical variables, the third column denotes the number of uniform bins used when they are converted to categorical. \emph{Basic} set: attributes 1--4 (2 numerical, 2 categorical; 14-dimensional distribution, 27-dimensional for discretized numerical). \emph{Socio} set: attributes 1--21 (8 numerical, 13 categorical; 68-dimensional distribution, 121-dimensional for discretized numerical). \emph{Extended} set: attributes 1--47 (21 numerical, 26 categorical; 230-dimensional distribution, 357-dimensional for discretized numerical).}
\label{tab:data}
\end{table}

The high dimensionality of the problem makes any use of the IPF algorithm impossible as the number of the corresponding matrix cells grows exponentially with the number of dimensions. For example, considering $n$ binary variables requires fitting of $2^n$ cells which quickly makes the algorithm impractical for large $n$.

\section{Results and Discussion}
\label{sec:results}

\subsection{Model specification and evaluation}
\label{sec:method:perform}

The three subsets of attributes (Basic, Socio and Extended) are used to run three separate experiments. Within each experiment, two separate cases are considered---when the numerical data are / are not converted to the categorical representation. For each case, we use 20\% of the data for model estimation (``training set'', approximately 0.5\% of the total population of Denmark) and the remaining 80\% for model evaluation (``test set'', also referred to as the ``true population'', approximately 2.0\% of the total population of Denmark). A pool of 100,000 agents is generated using each method. The evaluation is based on comparison of the statistical properties of the generated samples with the properties of the test set. The code for the methods used in the paper is available at \url{https://github.com/stasmix/popsynth}.

To optimize the hyper-parameters of the VAE, the training set for the VAE experiments is further subdivided into an actual training set (75\% of the original training set) and a validation set (25\% of the original training set). The new training set is used to fit the parameters of the VAE (weights and biases of the encoder and the decoder), while the validation set is used to compare performance of the models with different architectures and values of the hyper-parameters, which are optimized using grid search. For the encoder, we explore a fully connected ANN architecture consisting of 1, 2 or 3 hidden layers with 25, 50, 100; 50-25, 100-50; or 100-50-25 neurons, respectively. The decoder has the mirrored architecture of the encoder. The dimensionality of the latent space, $D_Z$, takes one of the following values: 5, 10, 25. For $\beta$, we use the values of 0.01, 0.05, 0.1, 0.5, 1.0, 10, 100. For the training, the RMSprop algorithm with the learning rate of 0.001 and $\rho=0.9$ is used, the size of a mini-batch is 64 and the number of epochs is 100. The numerical data are normalized before training to have zero mean and unit standard deviation. The best VAE model is selected according to the performance on the validation set for the four Basic variables, which is specified below.

For the Gibbs sampler (referred to as Gibbs-cond), the training data are used to calculate the full conditional distributions as frequency tables. During the sampling, the first $N_\mathrm{wu}=20,000$ samples are discarded and each successive $N_\mathrm{th}=20$ sample is kept, thus requiring 2,020,000 iterations to generate a pool of agents. 

For the BN, three different algorithms for the structure learning are used, which are implemented in the Pomegranate package for Python \citep{schreiber2018pomegranate}: Exact and greedy algorithms based on the MDL scoring with the A* heuristics, and the Chow-Liu tree building algorithm based on mutual information. We restrict the BN experiments to the case when numerical variables are converted to categorical, where all conditional probabilities are represented as frequency tables.

We also compare the statistical properties of the training set with the test set to verify that their distributions are similar. Additionally, we perform sampling from the marginal distribution of each variable, generating its value independently from the rest of the variables. This na\"ive baseline represents the extreme case when all the information about dependencies between the variables is missing and indicates the worst performance possible having the marginals perfectly reproduced.

To estimate how well the synthetic agents reproduce the statistical properties of the test data, we compare their empirical distributions $\hat{\pi}$ and $\pi$. The distributions are calculated as frequency tables with the numerical variables being discretized. More specifically, we calculate bin frequencies $\pi_{i\dots j}$ and $\hat{\pi}_{i\dots j}$ where each bin corresponds to a particular combination of values of the variables in the distribution. It is important to note that the comparison of high-dimensional distributions (and evaluation of generative models in general) represents an ongoing research challenge \citep{theis2015note}. The main problem is that the most common approaches based either on likelihood or kernel density estimation fail to provide reliable comparison due to the curse of dimensionality. To tackle this issue, instead of the direct comparison of the full joints, $\hat{\pi}(X_1, ..., X_n)$ and $\pi(X_1, ..., X_n)$, we use several different approaches which simplify the problem.

Following \cite{hu2018}, we compare the distributions being converted to the marginal distributions of all variables, $\pi(X_i)$, bivariate distributions of all possible pairs of variables, $\pi(X_i,X_j)$, and trivariate distributions of all possible triplets of variables, $\pi(X_i,X_j,X_k)$. We also compare projections of the distributions onto the 4-dimensional joint of the Basic attributes, $\pi(\mathrm{HousehNumPers},\mathrm{RespAgeCorrect},\mathrm{RespEdulevel},\mathrm{RespSex})$. To measure the difference between them, three standard metrics are calculated:
Standardized root mean squared error (SRMSE),
\begin{equation}
\label{eq:srmse}
\mathrm{SRMSE}(\hat{\pi}, \pi) = \frac{\mathrm{RMSE}(\hat{\pi}, \pi)}{\overline{\pi}}
= \frac{\sqrt{\sum_{i}\dots\sum_{j}\left(\hat{\pi}_{i\dots j}-\pi_{i\dots j}\right)^2/N_\mathrm{b}}}{\sum_{i}\dots\sum_{j}\pi_{i\dots j}/N_\mathrm{b}},
\end{equation}
where $N_\mathrm{b}$ is the total number of bins;
Pearson correlation coefficient (Corr); and coefficient of determination ($R^2$). The last two are defined in a standard way for the same bin frequencies. Other statistical metrics, such as the KL-divergence or the Wasserstein metrics, can be also used, however, are not considered in the paper. 

To compare correlation patterns in the distributions, we also perform a pairwise correlation analysis. We convert all the numerical variables to categorical for the sake of simplicity and calculate Cram\'er's V for all possible pairs of variables in the distributions. Additionally, the data is visually compared using the principal component analysis (PCA) described in \ref{sec:aapendixC}. 

To investigate the diversity of the generated samples, i.e. the ability of a model to generate out-of-sample agents instead of replicating the training data, we introduce the following procedure. For each sample generated by the model, $\hat{\mathbf{x}}_i$, we calculate its distance to all the samples used for training, $\mathbf{x}_j^\mathrm{train}$, as $\mathrm{RMSE}(\hat{\mathbf{x}}_i,\mathbf{x}_j^\mathrm{train})$ and find the nearest sample. Then, we calculate mean, $\mu_\mathrm{NS}$, and standard deviation, $\sigma_\mathrm{NS}$, of these nearest sample distances over all the generated samples. Thus, for the model which perfectly replicates the training data, $\mu_\mathrm{NS}=0$ and $\sigma_\mathrm{NS}=0$. For the sampling that uses only information from marginals and therefore does not capture statistical dependencies between the variables, these validation parameters tend to attain high values. However, it is difficult to interpret their values other than zero. 

\subsection{Comparison of methods}
\label{sec:results:microsamplecat}

Table~\ref{tab:results} summarizes the method evaluation results shown in Figs.~\ref{fig:scatter_1variate}, \ref{fig:scatter_2variate}, \ref{fig:scatter_3variate}, \ref{fig:scatter_farooq} and \ref{fig:scatter_pairwise}. The results verify that the training set naturally represents the best possible approximation to the true population given that it is the only information available. Its uniform expansion (re-sampling with replacement) produces the best results in terms of performance metrics, however, this approach is obviously incapable of generating new out-of-sample individuals. The population generated by the marginal sampler perfectly reproduces the marginals of the training set, however, it has a very high error for the rest of representations of the joint due to neglecting statistical dependencies between the attributes.

\begin{table}[t!]
\footnotesize
\centering
\begin{tabular}{l|ccccc|cc||ccccc|cc}
 & \multicolumn{7}{|c||}{Numerical converted to categorical} & \multicolumn{7}{c}{Numerical and categorical mixture} \\
\hline
Model & Marg. & Bivar. & Trivar. & Basic & Pair. & $\mu_\mathrm{NS}$ & $\sigma_\mathrm{NS}$ & Marg. & Bivar. & Trivar. & Basic & Pair. & $\mu_\mathrm{NS}$ & $\sigma_\mathrm{NS}$ \\
\hline
\hline
 & \multicolumn{7}{|c||}{Basic set (0-4/27)} & \multicolumn{7}{c}{Basic set (2-2/16)} \\
\hline
VAE & 0.069 & 0.167 & 0.309 & 0.482 & 0.083 & 0.001 & 0.017 & 0.144 & 0.289 & 0.481 & 0.696 & 0.239 & 0.032 & 0.090 \\
Gibbs-cond & \bf{0.017} & \bf{0.043} & \bf{0.102} & \bf{0.197} & \bf{0.026} & 0 & 0 & \bf{0.021} & \bf{0.048} & \bf{0.107} & \bf{0.201} & \bf{0.034} & 0.033 & 0.098 \\
BN-tree & 0.018 & 0.066 & 0.204 & 0.371 & 0.067 & 0.001 & 0.015 &  &  &  &  &  &  &  \\
BN-greedy & 0.023 & 0.063 & 0.142 & 0.278 & 0.082 & 0.001 & 0.014 &  &  &  &  &  &  &  \\
BN-exact & 0.024 & 0.065 & 0.157 & 0.308 & 0.051 & 0.001 & 0.015 &  &  &  &  &  &  &  \\
\hline
Marg. sampler & 0.021 & 0.481 & 1.098 & 1.740  & 1.213 & 0.037 & 0.094 & 0.021 & 0.481 & 1.098 & 1.740  & 1.213 & 0.037 & 0.094 \\
Training set & 0.018 & 0.042 & 0.094 & 0.182 & 0.035 & 0.002 & 0.048 & 0.018 & 0.042 & 0.094 & 0.182 & 0.035 & 0.002 & 0.048 \\
\hline
\hline
 & \multicolumn{7}{|c||}{Socio set (0-21/121)} & \multicolumn{7}{c}{Socio set (8-13/68)} \\
\hline
VAE & 0.095 & 0.244 & \bf{0.497} & 0.693 & \bf{0.234} & 0.004 & 0.035 & \bf{0.189} & \bf{0.435} & \bf{0.816} & \bf{0.847} & \bf{0.203} & 0.019 & 0.071 \\
Gibbs-cond & 0.851 & 2.087 & 4.080 & 4.025 & 1.063 & 0 & 0 & 1.523 & 3.680 & 7.192 & 5.224 & 1.187 & 0.008 & 0.045 \\
BN-tree & \bf{0.014} & \bf{0.209} & 0.552 & 1.211 & 0.464 & 0.013 & 0.058 &  &  &  &  &  &  &  \\
BN-greedy & 0.117 & 0.377 & 0.857 & \bf{0.571} & 0.466 & 0.001 & 0.018 &  &  &  &  &  &  &  \\
BN-exact & 0.123 & 0.339 & 0.729 & 0.836 & 0.366 & 0.004 & 0.033 &  &  &  &  &  &  &  \\
\hline
Marg. sampler & 0.016 & 0.550 & 1.438 & 1.740  & 1.246 & 0.037 & 0.094 & 0.016 & 0.550 & 1.438 & 1.740 & 1.246 & 0.037 & 0.094 \\
Training set & 0.013 & 0.032 & 0.069 & 0.182 & 0.097 & 0.002 & 0.048 & 0.013 & 0.032 & 0.069 & 0.182 & 0.097 & 0.002 & 0.048 \\
\hline
\hline
 & \multicolumn{7}{|c||}{Extended set (0-47/357)} & \multicolumn{7}{c}{Extended set (21-26/230)} \\
\hline
VAE & 0.155 & \bf{0.463} & \bf{0.989} & \bf{0.960} & \bf{0.317} & 0.003 & 0.030 & \bf{0.297} & \bf{0.807} & \bf{1.622} & \bf{1.184} & \bf{0.386} & 0.033 & 0.093 \\
Gibbs-cond & 2.045 & 6.216 & 14.200 & 27.275 & 1.428 & 0 & 0 & 2.337 & 6.746 & 14.897 & 27.330 & 1.428 & 0 & 0 \\
BN-tree & \bf{0.113} & 0.500 & 1.272 & 2.577 & 0.893 & 0.015 & 0.062 &  &  &  &  &  &  &  \\
\hline
Marg. sampler & 0.015 & 0.561 & 1.578 & 1.740 & 1.419 & 0.037 & 0.094 & 0.015 & 0.561 & 1.578 & 1.740 & 1.419 & 0.037 & 0.094 \\
Training set & 0.013 & 0.039 & 0.091 & 0.182 & 0.099 & 0.002 & 0.048 & 0.013 & 0.039 & 0.091 & 0.182 & 0.099 & 0.002 & 0.048 \\
\hline
\end{tabular}
\caption{
Standardized root mean squared error (SRMSE) between the synthetic agents generated using different methods (VAE, Gibbs-cond, and BN) and the test set for the different representations of the joint probability distribution: Marginal, bivariate, trivariate, a projection on the four Basic variables, and pairwise correlations measured by Cram\'er's V. Left/right parts of the table show the results when numerical variables are/are not converted to categorical, respectively. $\mu_\mathrm{NS}$ and $\sigma_\mathrm{NS}$ denote mean and standard deviation of the distances between each generated sample and its nearest sample in the training set. These two diversity measures for the training set are calculated w.r.t. the test set in contrast to the rest of the methods. The numbers in the parenthesis denote ``number of numerical variables''-``number of categorical variables''/``dimensionality of the one-hot encoded input''. See also Figs.~\ref{fig:scatter_1variate}, \ref{fig:scatter_2variate}, \ref{fig:scatter_3variate}, \ref{fig:scatter_farooq} and \ref{fig:scatter_pairwise} for more details.
}
\label{tab:results}
\end{table}

The results for the Gibbs sampler (Gibbs-cond) confirm its inability to generate new out-of-sample data when the sampling conditionals are estimated as frequency tables from the training data. The agents generated by the Gibbs sampler replicate the training samples when the numerical variables are converted to categorical ($\mu_\mathrm{NS}$ and $\sigma_\mathrm{NS}$ are exactly equal to zero). In the case when the mixture of both categorical and numerical variables is used, the Gibbs sampler introduces noise that depends on the size of the bins used to estimate the conditional probabilities. This fact explains its superior performance in the low-dimensional case when the Basic attributes are considered.  The Gibbs sampler is also prone to be stuck in a local minimum due to the curse of dimensionality. When the Socio case is considered, the Gibbs sampler becomes trapped in a local minimum around the starting point, generating a subset of agents from the training set. For the Extended case, the Gibbs sampler, which is initialized using an agent from the training set, cannot even escape the starting point. We also generated several chains from different starting points but it did not qualitatively change the overall behavior (results are not shown). To address this issue, various models to estimate the conditional probabilities, which are capable of dealing with sparse data, such as ANNs, can be explored.

\begin{figure}[t!]
\centering
(a)
\includegraphics[width=1.0\linewidth]{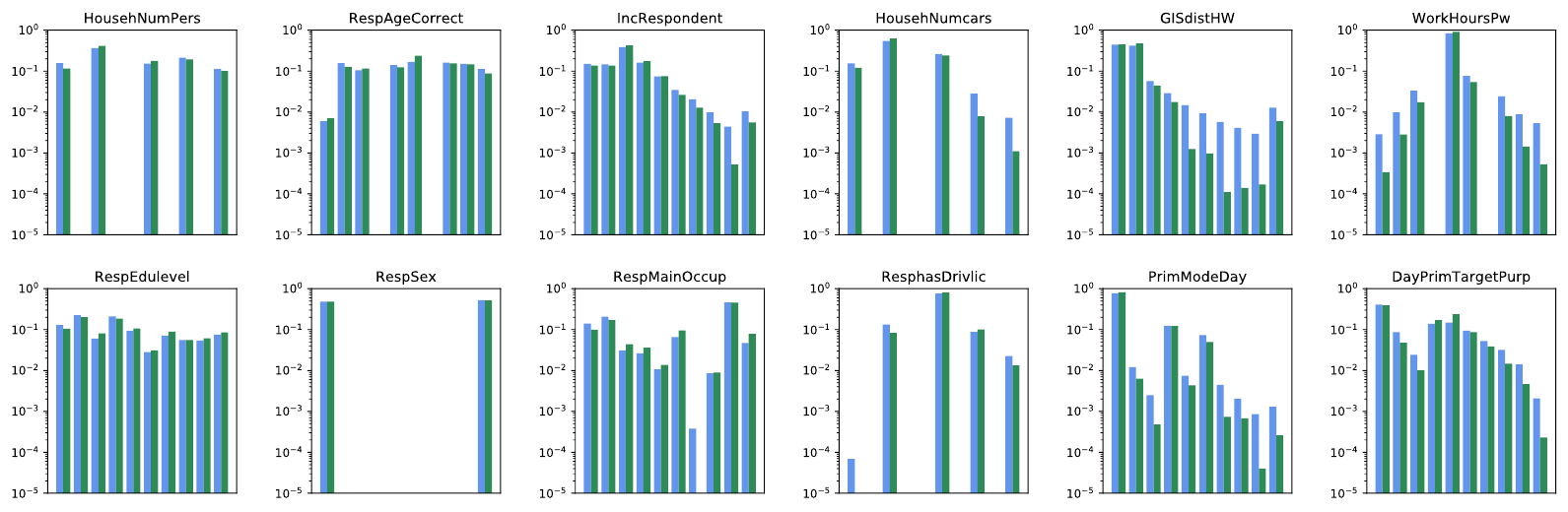}\\
(b)
\includegraphics[width=1.0\linewidth]{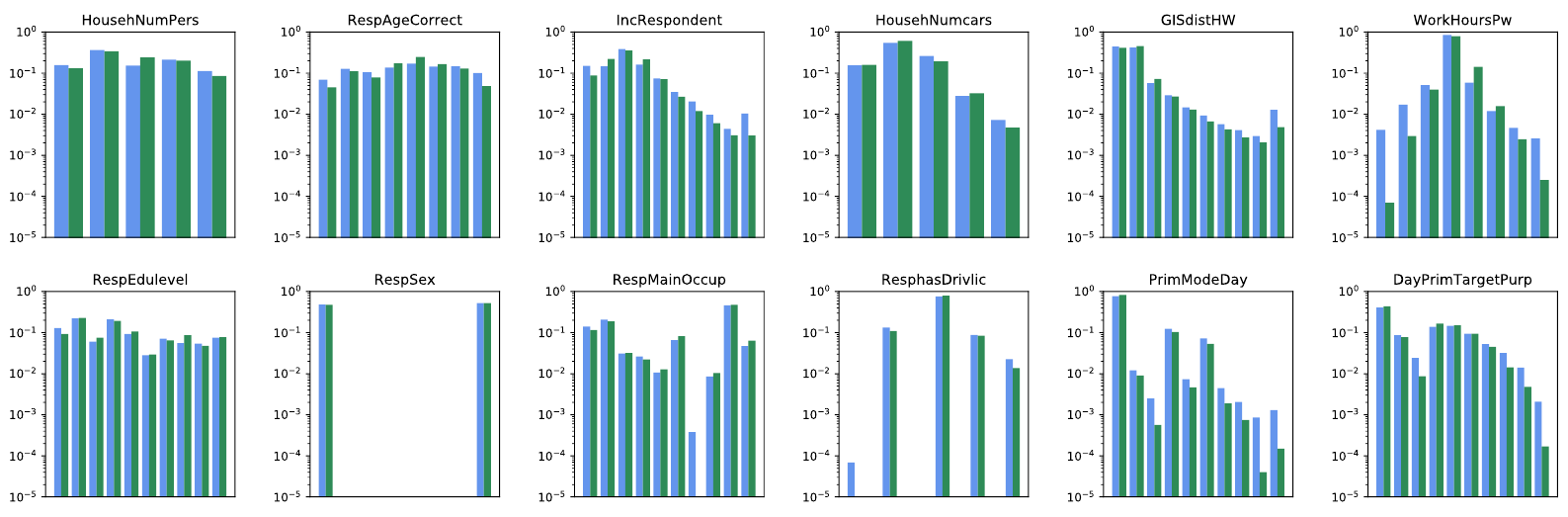}
\caption{
Examples of marginal distributions for the true population (blue) and samples generated using VAE (green) for the Extended attributes. (a) numerical variables are converted to categorical and (b) without the conversion. The first and second rows in both (a) and (b) contain numerical and categorical variables, respectively. The logarithmic scale is used to highlight low probability domains.}
\label{fig:marg}
\end{figure}

Although the VAE performance is acceptable for all the evaluation metrics considered, all the BN-based models outperform the VAE in the low-dimensional Basic case. However, its advantages become more evident when the number of modeled variables grows. The VAE performs on par with the BNs when the attributes from the Socio set are considered. It worth noting that the learning of the exact BN structure for 21 variables took almost 24 hours on a computer with 16 x 4.2~GHz CPU cores and 128~GB RAM. At the same time, the VAE training took around 10 minutes on the same machine using the GeForce GTX 1080 Ti GPU. For the Extended attributes, the VAE clearly outperforms the BN-tree model, while both exact and greedy BN structure learning algorithms did not converge after 24 hours of calculations. Figure~\ref{fig:marg} shows the VAE marginals for eight selected attributes from the Extended set. Although there are inconsistencies for the low probability values of some attributes (e.g., discretized GISdistHW in Fig.~\ref{fig:marg}a or WorkHoursPw in Fig.~\ref{fig:marg}b), the values that have high probability are captured correctly.

When the mixture of both numerical and categorical variables is used, VAE shows slightly worse performance compared to the case when the numerical variables are converted to categorical. This might be due to the coarse grid of hyper-parameters and the limited number of ANN architectures explored. Additionally, separate log-likelihood weighting factors of the numerical and categorical terms in the loss function (Eq.~(\ref{eq:loss})) can be  introduced. However, this approach can quickly become computationally expensive as the dimensionality of the optimization grid grows exponentially with the number of hyper-parameters involved. Nevertheless, as mentioned before, it is possible to avoid this problem by converting numerical variables to categorical using a sufficiently high resolution of the discretization bins.

Increasing the number of training epochs from 100 to 200 for the VAE trained on the Extended set slightly reduces the SMRSE for the Basic projection of the best performing model from 0.959 to 0.912 for discretized numerical variables and from 1.184 to 1.053 for non-discretized ones. This behavior suggests that the training stopped before reaching a minimum of the loss function (Eq.~(\ref{eq:loss})). Early stopping is known to be one of the regularization techniques used to prevent overfitting. However, the most important feature of VAE that helps to avoid this problem is its bottleneck structure. Indeed, the model is able to compress the data from initially sparse 357-dimensional space into the 25-dimensional latent space representation (Table~\ref{tab:vae_best}). The Basic and Socio attributes are compressed from 27 to 5 dimensions and from 121 to 10 dimensions respectively. Finally, all $\mu_\mathrm{NS}$ and $\sigma_\mathrm{NS}$ are non-zero for all the VAE models. Therefore, we can expect that VAE does not simply memorize the training data.

\begin{table}[ht!]
\centering
\begin{tabular}{l|cccc||cccc}
 & \multicolumn{4}{|c||}{Numerical converted to categorical} & \multicolumn{4}{c}{Numerical and categorical mixture} \\
\hline 
Attributes Set & $n$ & $D_Z$ & Encoder & $\beta$ & $n$ & $D_Z$ & Encoder & $\beta$ \\
\hline 
\hline 
Basic & 27 & 5 & 100-50-25 & 0.5 & 16 & 10 & 100-50 & 0.1 \\
Socio & 121 & 25 & 100 & 0.5 & 68 & 10 & 100-50-25 & 0.5 \\
Extended & 357 & 25 & 100 & 0.5 & 230 & 25 & 100-50 & 0.5 \\
\hline
\end{tabular}
\caption{The best VAE architectures selected using grid search. Left/right parts of the table show the results when numerical variables are/are not converted to categorical, respectively. Columns: set of attributes, dimensionality of the data space ($n$), dimensionality of the latent space ($D_Z$), architecture of the decoding ANN, and regularization strength ($\beta$). ANN architecture is denoted as ``number of neurons in hidden layer 1''-``number of neurons in hidden layer 2''-\dots The architecture of the decoder is a mirror of the encoder.}
\label{tab:vae_best}
\end{table}

\begin{figure}[ht!]
\centering
(a)
\includegraphics[width=1.0\linewidth]{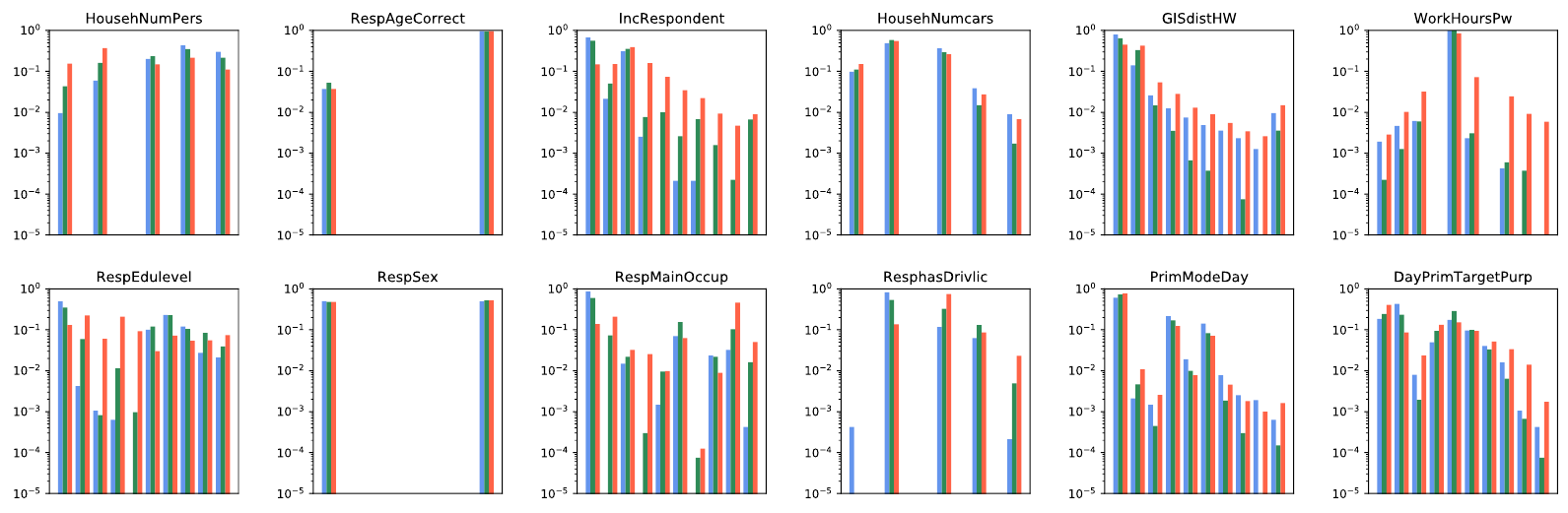}\\
(b)
\includegraphics[width=1.0\linewidth]{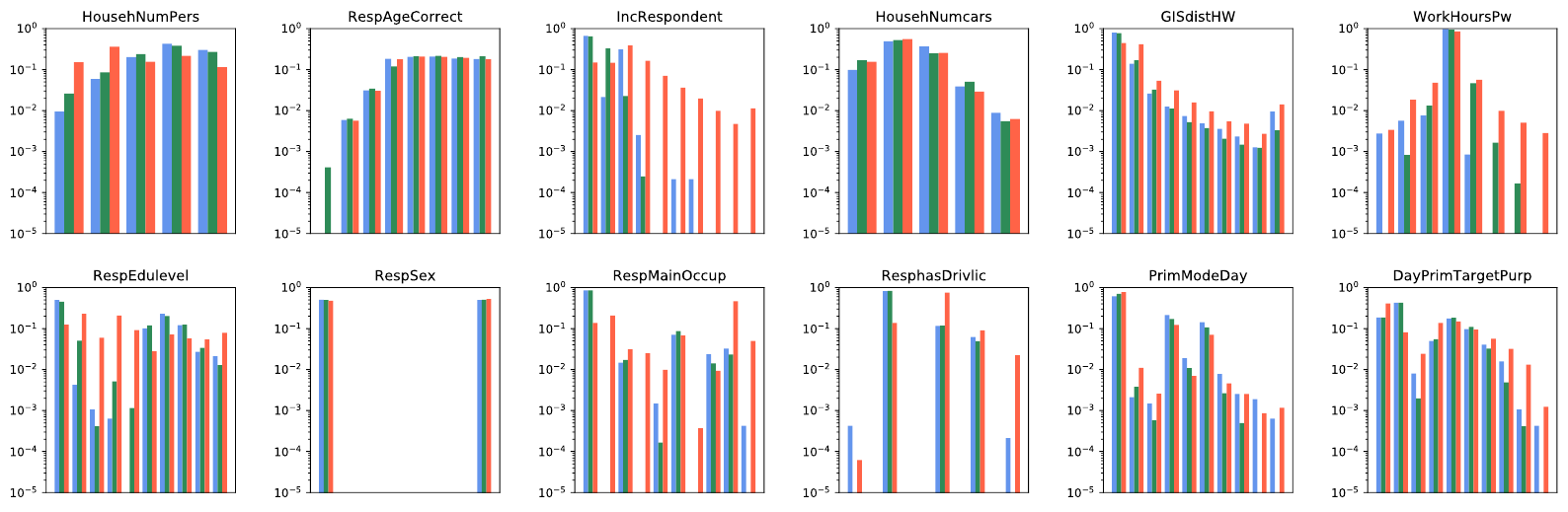}
\caption{
Examples of marginal distributions for the true population (blue), samples generated using VAE (green) and Gibbs sampler for marginals (red) for the Extended attributes for the population under 20 years old. (a) numerical variables are converted to categorical and (b) without the conversion. The first and second rows in both (a) and (b) contain numerical and categorical variables, respectively. The logarithmic scale is used to highlight low probability domains.}
\label{fig:marg_u20}
\end{figure}

In contrast to the popular deep generative modeling applications such as image generation, it is more difficult to conclude whether the generated out-of-sample agents represent plausible individuals (sampling zeros) or not (structural zeros). One of the possible ways is to test the synthetic population for logical inconsistencies using a predefined set of rules. For instance, the inconsistencies could be classified as children with university degrees or unemployed individuals with non-zero working hours. As an example, the same subset of marginal distributions from Fig.~\ref{fig:marg} is plotted for the individuals who are under 20 years old in Fig.~\ref{fig:marg_u20}. These distributions are quite different from the whole population, especially for personal income (IncRespondent), educational attainment (RespEdulevel), main occupation (RespMainOccup) and possession of a driving license (ResphasDrivlic). The VAE model correctly reproduces the observed pattern apart from some low probability domains. 

While it is feasible to inspect such pairwise correlations based on handcrafted rules for a small set of attributes, this approach becomes impracticable for a large number of attributes, especially taking into account high-order correlations (e.g. when an employed person who lives close to the place of occupation has long commuting time). Although there is no silver bullet solution to differentiate between sampling and structural zeros, the most crucial rules for application purposes can be either implemented during the re-sampling stage or directly incorporated into the model \citep{hu2018}. On the other hand, the synthesizers that repeatedly sample from a micro-sample, replicating the original agents, will never have this problem. This trade-off between overfitting and generalization remains an important conceptual question in generative modeling.

\section{Conclusions and future work}
\label{sec:conlude}

This paper presents a scalable approach to the problem of generating large sample pools of agents for use in micro-simulations of transport systems. The approach is based on deep generative modeling framework and is capable of generating synthetic populations with potentially hundreds of mixed continuous and categorical attributes. This opens a path toward modeling of a population with a very high level of detail, including smaller zones, personal details and travel preferences.

The proposed approach is based on unsupervised learning of the joint distribution of the data using a Variational Autoencoder (VAE) model. The model uses a deep artificial neural network that transforms the initially sparse data into a compressed latent space representation which allows for efficient sampling. This is however not a unique feature of the VAE and other latent variable models can be also considered to deal with sparse data.

The presented study case involves the synthesis of agents from a large Danish travel diary, where around 29,000 records about its participants are used as training data and around 117,000 records are used as test data. It is shown that while the other two common population methods based on the Gibbs sampler and Bayesian Networks (BNs) outperform the VAE for small-scale problems, the advantages of the VAE become evident when the number of attributes increases. The VAE and BN also allow for growing ``out-of-sample'' agents which are different from the agents present in the training data but preserve the same statistical properties and dependencies as in the original data. Such an approach can be used to address the sampling zeros problem. This is contrary to the Gibbs sampler that replicates the agents from the original sample when the underlying conditional distributions are approximated as multidimensional frequency tables.

This ability of generative models to generate out-of-sample agents can be used to explore a richer set of modeling scenarios. It also addresses data privacy issues since these agents do not directly correspond to real people anymore. This feature can facilitate creation of publicly open datasets by fitting generative models to the data containing private information. These synthetic datasets can be used, for example, to test models that were initially tailored for specific cases on a wider range of scenarios. 

\subsection{Future work}

In this paper, the most basic VAE architecture is considered. The most straightforward extension of the proposed model is the conditional VAE \citep{NIPS2015_5775}, which estimates probability distributions conditional on other variables. For instance, it can be used to address the job formation problem where the job distribution is conditional on the underlying population. Numerous other variations of VAEs can also be explored, such as discrete latent space VAE \citep{jang2016categorical}, Wasserstein VAE \citep{tolstikhin2017wasserstein} or Ladder VAE \citep{NIPS2016_6275} to name a few. The VAE model can also incorporate a recurrent neural network architecture to generate sequential data, such as fine-grained daily activity chains including time, locations, transportation modes and purpose. Finally, other deep generative models which make use of compact latent space representations, such as Generative Adversarial Networks, Restricted Boltzmann Machines or Deep Belief Networks, can be explored. For example, generative adversarial networks can potentially produce agents with less logical inconsistencies. They are, however, much more difficult to train in practice.

In the current work, only the first stage of simulation-based approaches related to growing pools of micro-agents is addressed. The next important stage involves re-sampling, where individuals are selected from the simulated pool in such a way that the resulting population is aligned with imposed properties of future populations. In this way, changes in income and age composition can be reflected. Another challenge that can be addressed at the re-sampling stage is the explicit elimination of illogical or inconsistently generated agents. This may involve the elimination of children with full-time work or high income. For the generation of complete trip diaries, this challenge becomes non-trivial, because the spatio-temporal ordering of the activities needs to be preserved.

\section*{Acknowledgements}

The research leading to these results has received funding from the European Union's Horizon 2020 research and innovation programme under the Marie Sklodowska-Curie grant agreement No.~713683 (COFUNDfellowsDTU), and from the European Union's Horizon 2020 research and innovation programme under the Marie Sklodowska-Curie Individual Fellowship H2020-MSCA-IF-2016, ID number 745673. The authors also thank the anonymous referees for useful comments.


\appendix


\section{Toy problem}
\label{sec:aapendixA}

We describe a toy problem to demonstrate the methodologies used in the paper in a more detailed fashion. We assume a population of agents characterized by two binary attributes, $X$ and $Y$, which values can be either 0 or 1 (e.g., gender and car ownership). We also assume that the whole population consists of equal proportions of the two prototypical agents, $s_{0}\equiv(X=0, Y=0)$ and $s_{1}\equiv(X=1, Y=1)$ (e.g., corresponding to a woman without a car and a man with a car), while the two other combinations of the attributes are not present. This population has the joint probability distribution $P(X,Y)$ as shown in Table~\ref{tab:appendix:toy:gibbs}a.

\begin{table}[ht!]
    \begin{subtable}{.3\linewidth}
      \centering
        \caption{$P(X,Y)$}
        \begin{tabular}{c|cc}
            \backslashbox{X}{Y} & 0 & 1 \\
            \hline
            0 & 0.5 & 0 \\
            1 & 0 & 0.5 \\
        \end{tabular}
    \end{subtable}%
    \begin{subtable}{.3\linewidth}
      \centering
        \caption{$P(X|Y)$}
        \begin{tabular}{c|cc}
            \backslashbox{X}{Y} & 0 & 1 \\
            \hline
            0 & 1 & 0 \\
            1 & 0 & 1 \\
        \end{tabular}
    \end{subtable} 
    \begin{subtable}{.3\linewidth}
      \centering
        \caption{$P(Y|X)$}
        \begin{tabular}{c|cc}
            \backslashbox{X}{Y} & 0 & 1 \\
            \hline
            0 & 1 & 0 \\
            1 & 0 & 1 \\
        \end{tabular}
    \end{subtable} 
\caption{Joint (a) and conditional (b,c) probability distributions for the toy problem.}
\label{tab:appendix:toy:gibbs}
\end{table}

\subsection{Gibbs sampler}
The Gibbs sampler requires conditional probabilities for sampling. In the paper, we consider the simplest case when the conditional probabilities are estimated as frequency tables. For the toy problem presented here, these conditional probabilities $P(X|Y)$ and $P(X|Y)$ are shown in Table~\ref{tab:appendix:toy:gibbs}a and Table~\ref{tab:appendix:toy:gibbs}b, respectively. As can be immediately seen from them, when the sampling starts at the point $(X=0, Y=0)$, the probability of jumping to another state $(X=1, Y=1)$ is zero. Thus, the sampler becomes trapped in this starting point. Moreover, the number of such probability islands can grow exponentially due to the curse of the dimensionality. While it is possible to cover different starting points in low dimensions, the number of possible combinations grows exponentially with the number of variables involved and the number of the values they can take. Finally, as the probabilities of other combinations of the attributes are exactly zero, the Gibbs sampler will always overfit the original data unless priors for the sampling zeros are imposed.

\subsection{Variational Autoencoder}
\emph{Model specification.} For the toy problem, we assume the simplest VAE architecture, where both the encoder and the decoder are represented by linear transformations and the dimensionality of the latent variable is one ($D_Z=1$). Because $X$ and $Y$ are binary, we do not use the one-hot representation. The outline below can be can be trivially reformulated for one-hot encoded variables. The encoder has two inputs, $x$ and $y$, corresponding to the values of $X$ and $Y$. For each sample $s\equiv(x,y)$, the encoder outputs two vectors, corresponding to the mean and logarithm of the standard deviation for the variable $Z$, calculated as $\mu(s) = w_{11}^\mathrm{e}x+w_{12}^\mathrm{e}y+b_1^\mathrm{e}$ and $\log\sigma(s) = w_{21}^\mathrm{e}x+w_{22}^\mathrm{e}y+b_2^\mathrm{e}$. Thus, the encoder has six parameters, $\theta\equiv\{ w_{11}^\mathrm{e}, w_{12}^\mathrm{e}, b_1^\mathrm{e}, w_{21}^\mathrm{e}, w_{22}^\mathrm{e}, b_2^\mathrm{e} \}$. The latent variable is calculated as $z(s)=\mathcal{N}(\mu, \sigma) = w_{11}x+w_{12}y+b_1 + \exp\left(w_{21}x+w_{22}y+b_2\right)\epsilon$, where $\epsilon$ is sampled from $\mathcal{N}(0, 1)$ for every evaluation of the encoder output. The decoder has just the logistic regression output for each sample $\hat{s}\equiv(\hat{x},\hat{y})$, $\hat{x} = \left[1 + \exp\left(-w_{1}^\mathrm{d}z-b_1^\mathrm{d}\right)\right]^{-1}$, $\hat{y} = \left[1 + \exp\left(-w_{2}^\mathrm{d}z-b_2^\mathrm{d}\right)\right]^{-1}$. Thus, the decoder has four parameters, $\phi\equiv\{ w_{1}^\mathrm{d}, b_1^\mathrm{d}, w_{2}^\mathrm{d}, b_2^\mathrm{d} \}$.

\begin{figure}[ht!]
\centering
\includegraphics[height=5.0cm]{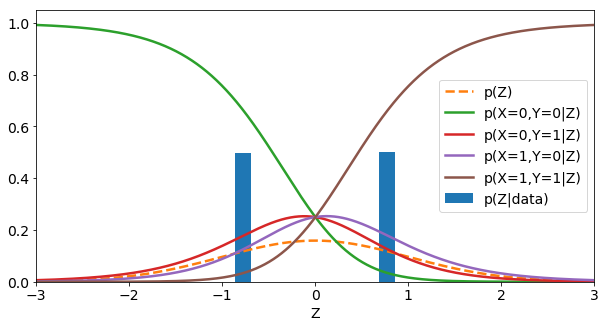}
\caption{The latent space of the VAE trained for the toy problem.}
\label{fig:appendix:toy:vae}
\end{figure}

\emph{Training phase.} The objective of the training phase is to minimize the loss function (Eq.~(\ref{eq:loss})) w.r.t. the model parameters $\theta$ and $\phi$ over the training set consisting of the two tanning samples $s_0$ and $s_1$. For each optimization step, the loss function is calculated as a sum of losses for these two tanning samples, and its gradients w.r.t. the model parameters are estimated numerically. For each tanning sample $s_k$ during each optimization step, the model calculates corresponding $\mu_k\equiv\mu(s_k)$ and $\log\sigma_k\equiv\log\sigma(s_k)$, randomly samples $z_k$ using $\mu_k$ and $\sigma_k$, and outputs probabilities of $X$ and $Y$ as the components of $\hat{s}_k$. Then, the two error terms in the loss function are calculated. The reconstruction error term (Eq.~(\ref{eq:loss_cat})) estimates the difference between the input $s_k$ and the output $\hat{s}_k$ vectors. For example, for the prototypical individual $s_0$, it will adjust $\theta$ and $\phi$ so both $\hat{x}$ and $\hat{y}$ become closer to 0. This error term will effectively push $\mu_0$ and $\mu_1$ apart and try to decrease $\log\sigma_0$ and $\log\sigma_1$ to minus infinity. At the same time, the KL-divergence error term (Eq.~(\ref{eq:loss_KL})) will try to push $\mu_0$, $\mu_1$, $\log\sigma_0$, and $\log\sigma_1$ to zero. These two error terms counterbalance each other assuring that (i) the samples are properly reconstructed and (ii) the latent space has a probabilistic meaning. For example, after 1000 optimization steps, $\mu$ and $\log\sigma$ converge to the following values: $\mu_0 = -0.891$, $\mu_1 = 0.889$, $\log\sigma_0 = -1.008$, $\log\sigma_0 = -1.003$. It means that the prototypical samples are encoded (i.e., mapped using the encoder) to the latent space as these mean values (see $P(Z|\mathrm{data})$ in Fig.~\ref{fig:appendix:toy:vae}).

\emph{Sampling phase.} During the sampling stage, the encoder is discarded and only the decoder is used. To generate samples from the trained model, $z$ is sampled from the prior $\mathcal{N}(0, 1)$ and mapped through the decoder to the probabilities $\hat{x}$ and $\hat{y}$. The final value for each component is assumed to be 0 if its probability is less 0.5 and 1 otherwise. The decoder output probabilities depicted in Fig.~\ref{fig:appendix:toy:vae} show that the model will generate $s_0$ if $z<0$ and $s_1$ if $z>0$. For the toy problem, the model cannot generate out-of-sample agents because of the perfect separation of the training samples in the latent space. However, for more complex problems, some regions of the latent space will be mapped to out-of-sample agents. This property is related to the information bottleneck caused by the dimensionality of the latent space and the prior used.

\subsection{Bayesian network}
\begin{figure}[ht!]
\centering
\includegraphics[height=2.0cm]{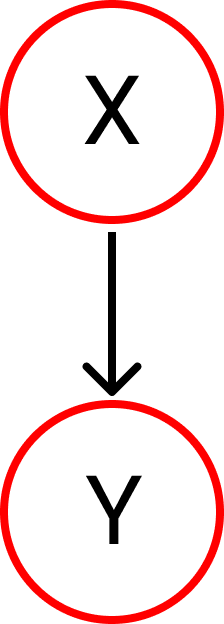}
\caption{The BN sructure learned for the toy problem.}
\label{fig:appendix:toy:bn}
\end{figure}

The joint distribution of the two variables $P(X,Y)$ can be factorized in three different ways: $P(X)P(X|Y)$ corresponding to the graph $X \rightarrow Y$, $P(Y)P(Y|X)$ corresponding to the graph $Y \rightarrow X$, and $P(Y)P(X)$ corresponding to the disconnected graph. From a statistical point of view, the first two representations are equivalent unless the causality problem is concerned. The disconnected BN can be represented as two marginals, $P(X=0)=P(X=1)=0.5$ and $P(Y=0)=P(Y=1)=0.5$. The connected BN can be represented as one marginal $P(X=0)=P(X=1)=0.5$ and one conditional $P(Y|X)$ shown in Table~\ref{tab:appendix:toy:gibbs}c. The MDL score function consists of the two terms, the data log-likelihood and the DAG model complexity which is subtracted from the former. Although the disconnected BN has lower complexity than the connected BN, its log-likelihood is lower than the gain in model complexity. As a result, the connected BN (Fig.~\ref{fig:appendix:toy:bn}) will have a higher score.

\section{Comparison of the distributions}
\label{sec:aapendixB}

\begin{figure}[ht!]
\centering
\includegraphics[width=0.95\linewidth]{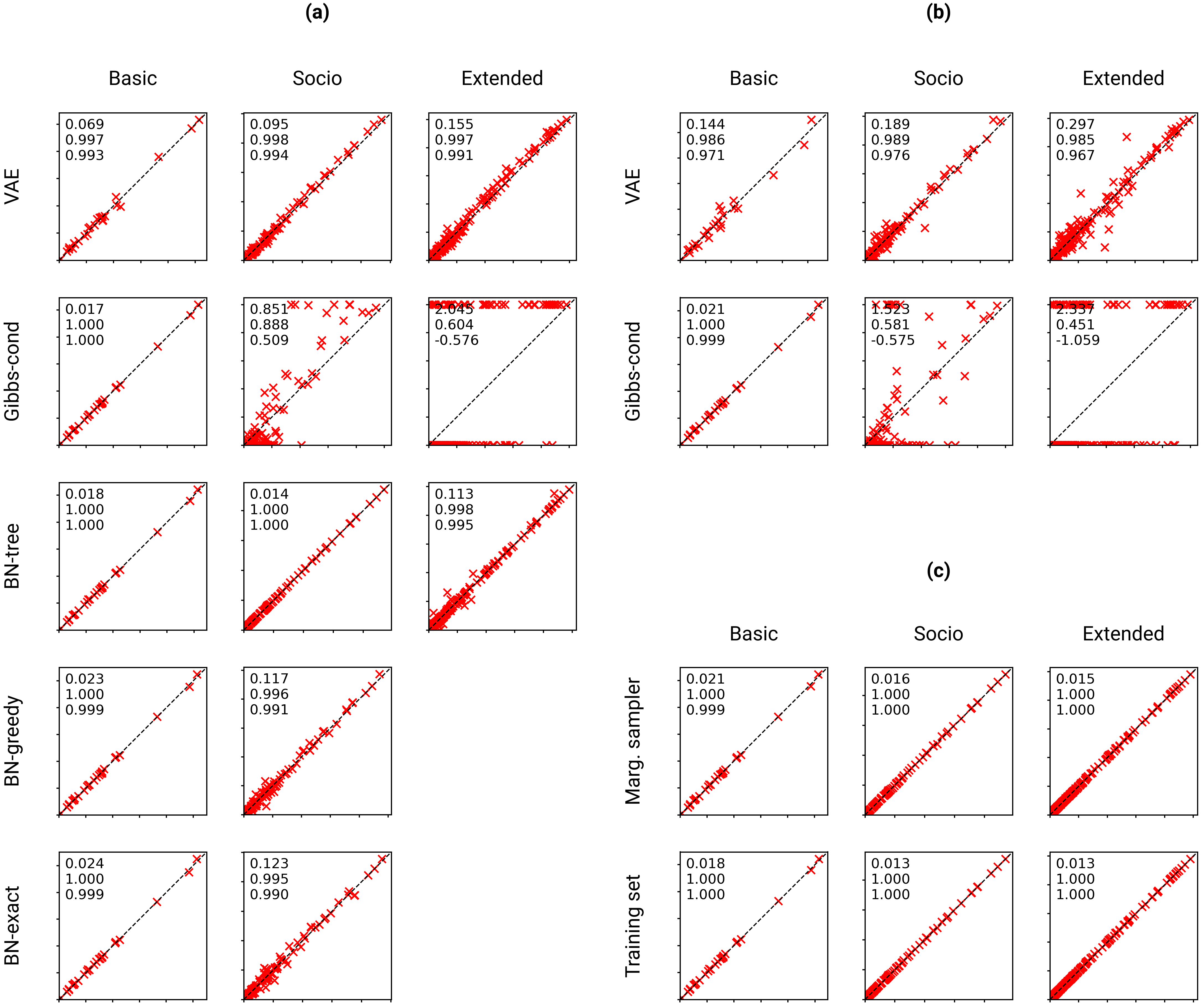}
\caption{Comparison of the marginal distributions for the test set (horizontal axis) and the synthetic agents generated using different methods (vertical axis). Each cross represents frequency of each value of each variable. The diagonal line corresponds to the perfect agreement. The three numbers correspond to the standardized root mean squared error (SRMSE), Pearson’s correlation coefficient (Corr) and coefficient of determination ($R^2$). (a) numerical variables are converted to categorical, (b) numerical variables are not converted to categorical, (c) baselines, which performance is independent of the conversion. The results are summarized in Table~\ref{tab:results}.}
\label{fig:scatter_1variate}
\end{figure}

\begin{figure}[ht!]
\centering
\includegraphics[width=0.95\linewidth]{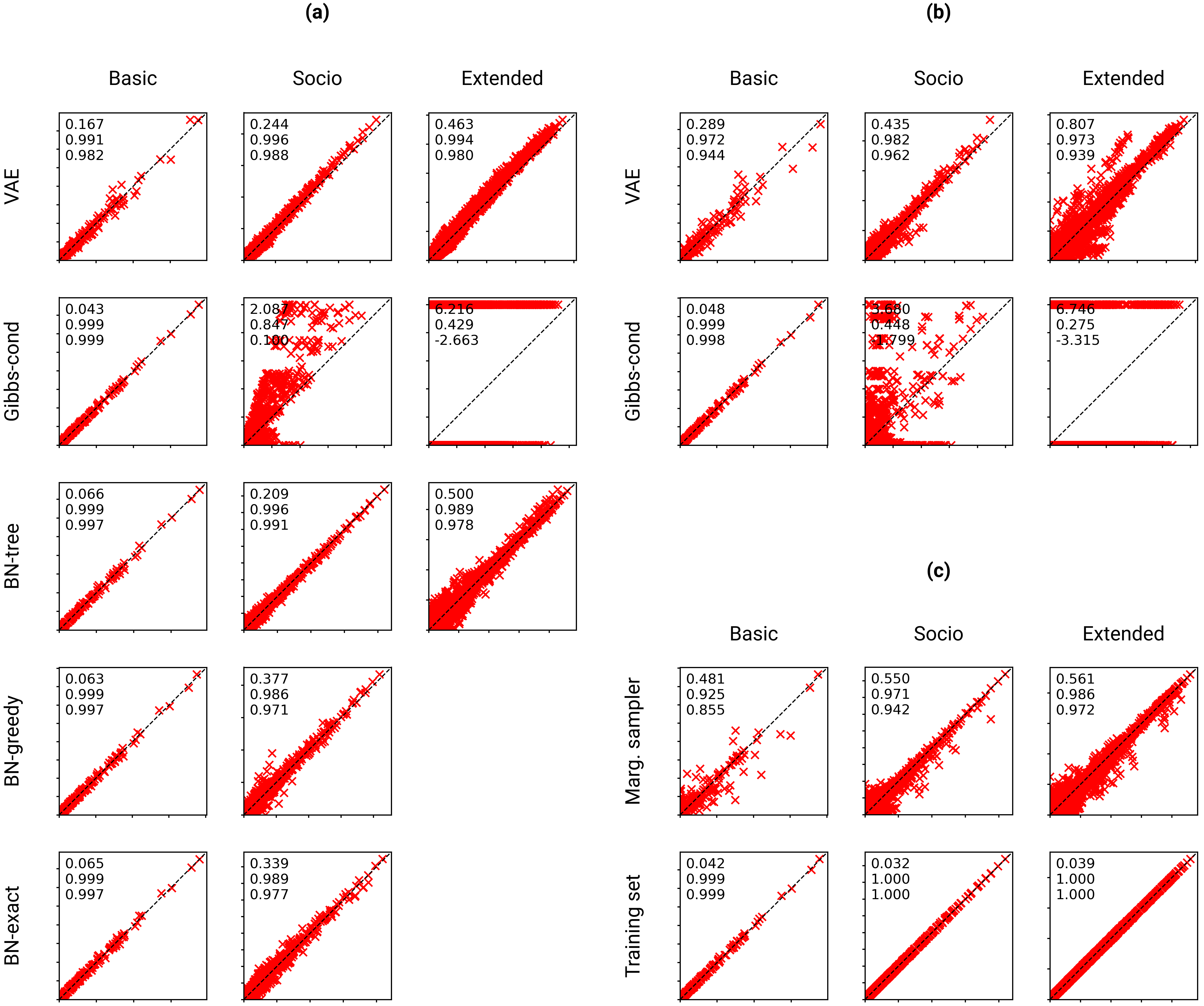}
\caption{Comparison of the bivariate distributions for the test set (horizontal axis) and the synthetic agents generated using different methods (vertical axis). Each cross represents frequency of each combination of values of every possible pair of variables. The diagonal line corresponds to the perfect agreement. The three numbers correspond to the standardized root mean squared error (SRMSE), Pearson’s correlation coefficient (Corr) and coefficient of determination ($R^2$). (a) numerical variables are converted to categorical, (b) numerical variables are not converted to categorical, (c) baselines, which performance is independent of the conversion. The results are summarized in Table~\ref{tab:results}.}
\label{fig:scatter_2variate}
\end{figure}

\begin{figure}[ht!]
\centering
\includegraphics[width=0.95\linewidth]{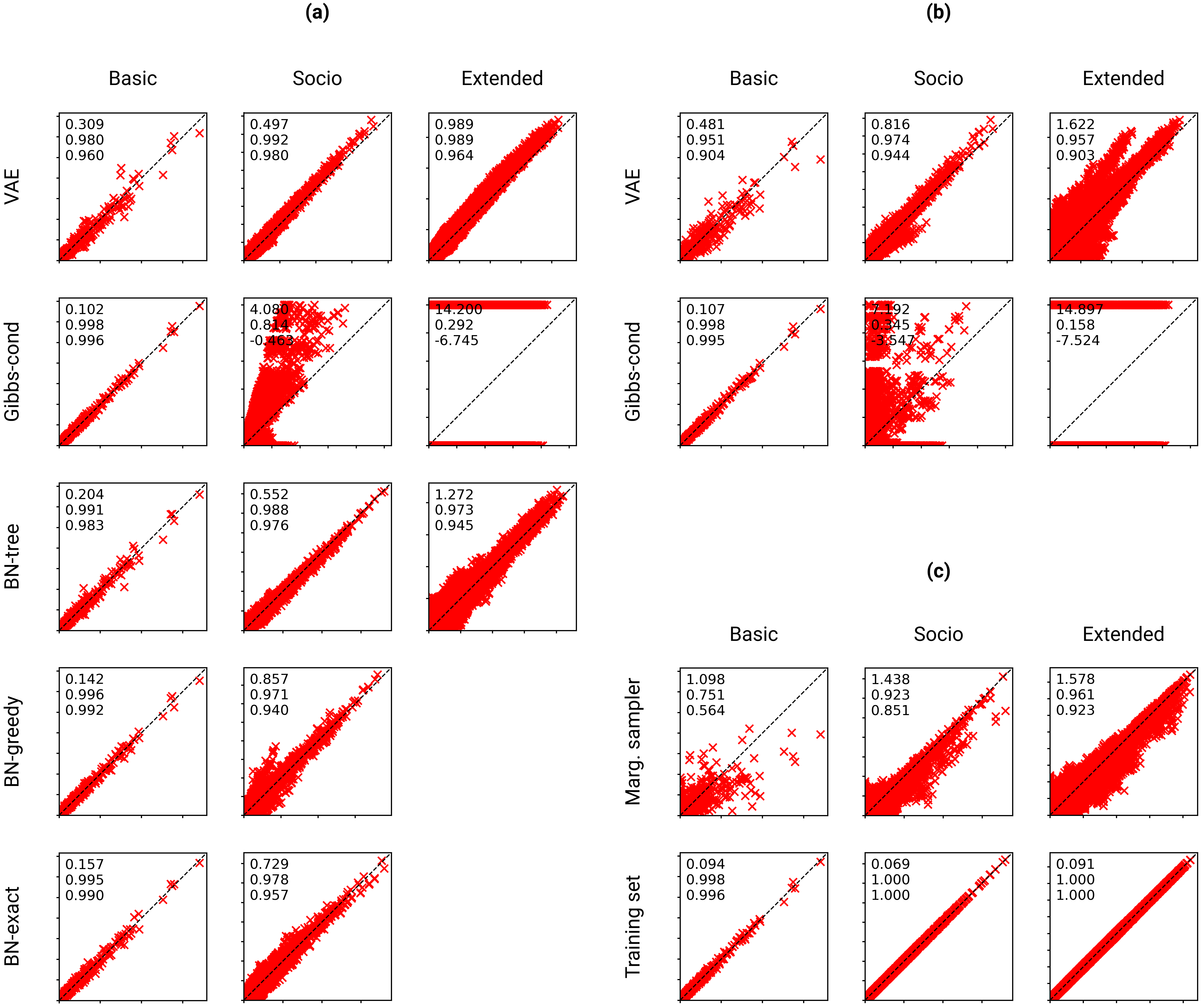}
\caption{Comparison of the trivariate distributions for the test set (horizontal axis) and the synthetic agents generated using different methods (vertical axis). Each cross represents frequency of each combination of values of every possible triplet of variables. The diagonal line corresponds to the perfect agreement. The three numbers correspond to the standardized root mean squared error (SRMSE), Pearson’s correlation coefficient (Corr) and coefficient of determination ($R^2$). (a) numerical variables are converted to categorical, (b) numerical variables are not converted to categorical, (c) baselines, which performance is independent of the conversion. The results are summarized in Table~\ref{tab:results}.}
\label{fig:scatter_3variate}
\end{figure}

\begin{figure}[ht!]
\centering
\includegraphics[width=0.95\linewidth]{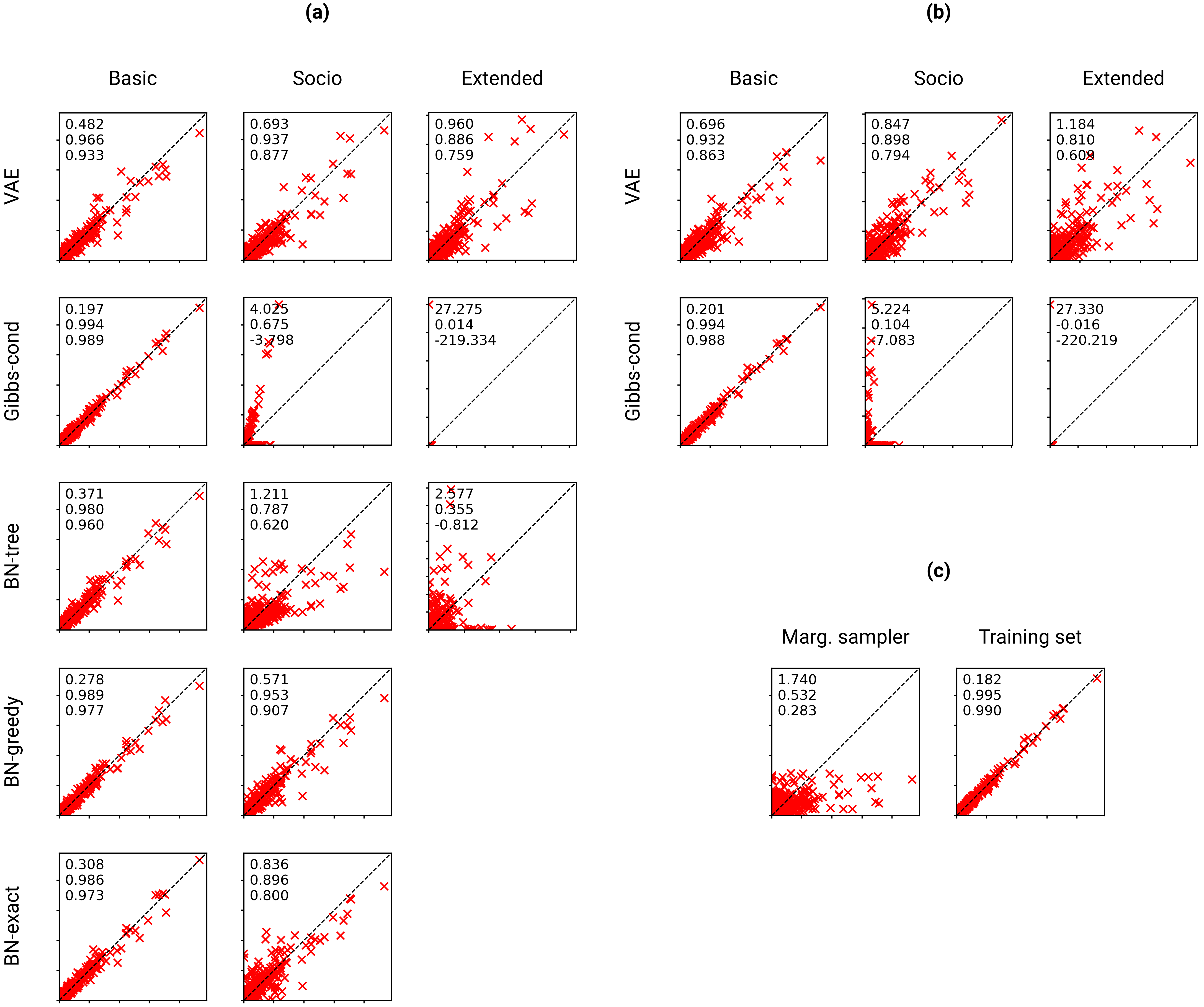}
\caption{Comparison of the joint distributions projected on the four Basic variables for the test set (horizontal axis) and the synthetic agents generated using different methods (vertical axis). Each cross represents frequency of each combination of values of the four Basic variables, where the other variables are marginalized. The diagonal line corresponds to the perfect agreement. The three numbers correspond to the standardized root mean squared error (SRMSE), Pearson’s correlation coefficient (Corr) and coefficient of determination ($R^2$). (a) numerical variables are converted to categorical, (b) numerical variables are not converted to categorical, (c) baselines, which performance is independent of the conversion. The results are summarized in Table~\ref{tab:results}.}
\label{fig:scatter_farooq}
\end{figure}

\begin{figure}[ht!]
\centering
\includegraphics[width=0.95\linewidth]{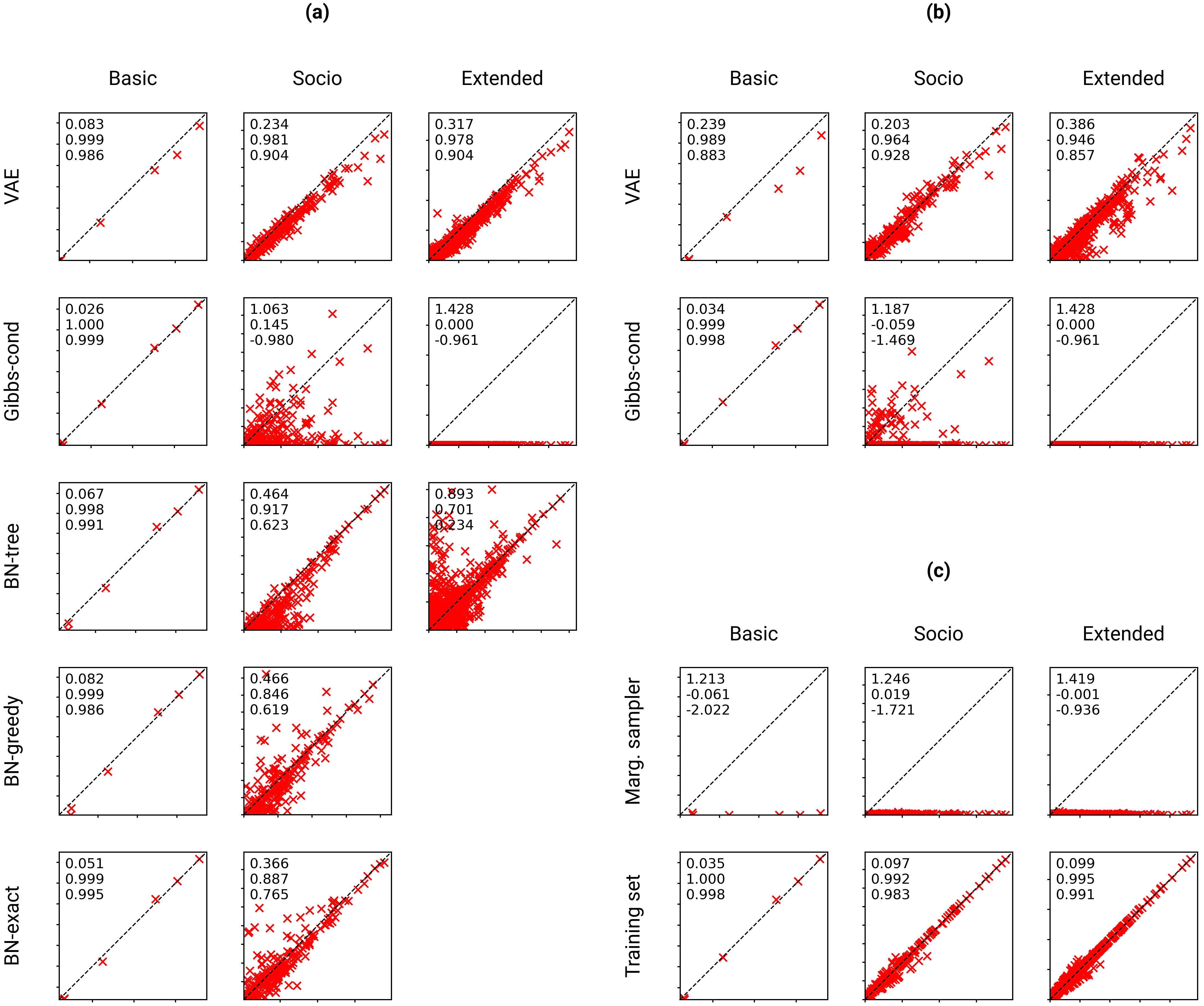}
\caption{Comparison of the pairwise correlation for the test set (horizontal axis) and the synthetic agents generated using different methods (vertical axis). Each cross represents the Cram\'er's V value for every possible pair of variables. The diagonal line corresponds to the perfect agreement. The three numbers correspond to the standardized root mean squared error (SRMSE), Pearson’s correlation coefficient (Corr) and coefficient of determination ($R^2$). (a) numerical variables are converted to categorical, (b) numerical variables are not converted to categorical, (c) baselines, which performance is independent of the conversion. The results are summarized in Table~\ref{tab:results}.}
\label{fig:scatter_pairwise}
\end{figure}

In this appendix, we show scatter plots illustrating performance of the models. In Figs.~\ref{fig:scatter_1variate}, \ref{fig:scatter_2variate} and \ref{fig:scatter_3variate}, we compare marginal (bin frequencies for single variables compared all together), bivariate (bin frequencies for every possible pair of variables compared all together), and trivariate (bin frequencies for every possible triplet of variables compared all together) distributions for the test set and the synthetic pools of agents generated using different methods. In Fig.~\ref{fig:scatter_farooq}, the full joint distributions are projected to the joints of the four Basic variables, where the rest of the variables are marginalized in the Socio and Extended cases. Finally, the pairwise correlations measured by Cram\'er's V for each possible pair of variables are compared in Fig.~\ref{fig:scatter_pairwise}.

\section{Principal component analysis}
\label{sec:aapendixC}

As the direct comparison of multidimensional distributions is difficult in high dimensions, one of the possible ways to simplify the problem is to reduce their dimensionality first. In Section~\ref{sec:results} and \ref{sec:aapendixB}, this reduction is based on marginalizing different subsets of the variables. An alternative approach can involve other dimensionality reduction methods such as Principal Component Analysis (PCA), Multi-dimensional Scaling (MDS) or t-distributed Stochastic Neighbor Embedding (t-SNE) to name but a few.

PCA is a method to find a transformation of data into a set of orthogonal principal components (PCs), where each PC represents a linear combination of the original variables. PCs are ordered in such way that the first PC corresponds to the direction of the highest possible variance and each succeeding PC has the highest variance possible given that it is orthogonal to the preceding PCs. PCA is widely used for visualization of high-dimensional data, which helps to reveal patterns. For example, when PCA was performed for the DNA sequences of Europeans \citep{pca_dna}, the data projected on the first PC plotted against the second PC closely resembled the map of Europe, meaning that the variations in the genes largely can be explained by a geographic location. In such plots, when PCs are plotted against each other, data points which are similar tend to be close to each other forming clustering patterns characteristic to the data. These plots can be also used for visual comparison of different datasets. 

For each of the attribute sets, we perform PCA for the training data, where numerical variables are converted to categorical and the one-hot encoded representation of categorical variables is used. Then, we project the training data together with 10,000 agents generated by the VAE and the BN onto different pairs of the PCs of the training data to visually compare their clustering patterns. The data points are aggregated to histograms for the data projections onto a single PC.

\begin{figure}[ht!]
\centering
\includegraphics[width=0.9\linewidth]{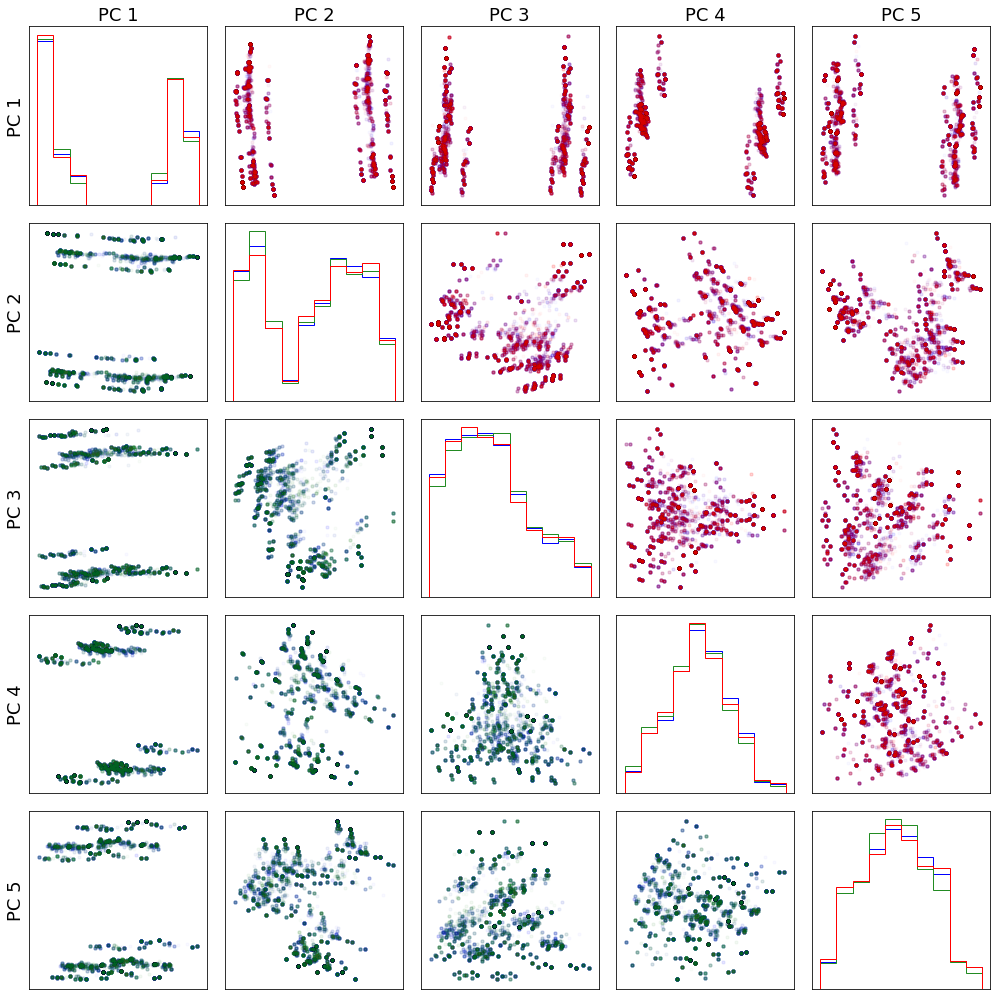}
\caption{Principal component analysis for the Basic case (numerical variables are converted to categorical). The data are projected to the first 5 principal components (PCs) of the training set (blue) and synthetic populations produced by VAE (green) and BN-exact (red). Diagonal plots show histograms of the data projected on the corresponding PC. On the off-diagonal plots, the PC projections are plotted one against each other, where each dot represents a data point projected on the two PCc.
}
\label{fig:pca1}
\end{figure}

\begin{figure}[ht!]
\centering
\includegraphics[width=0.9\linewidth]{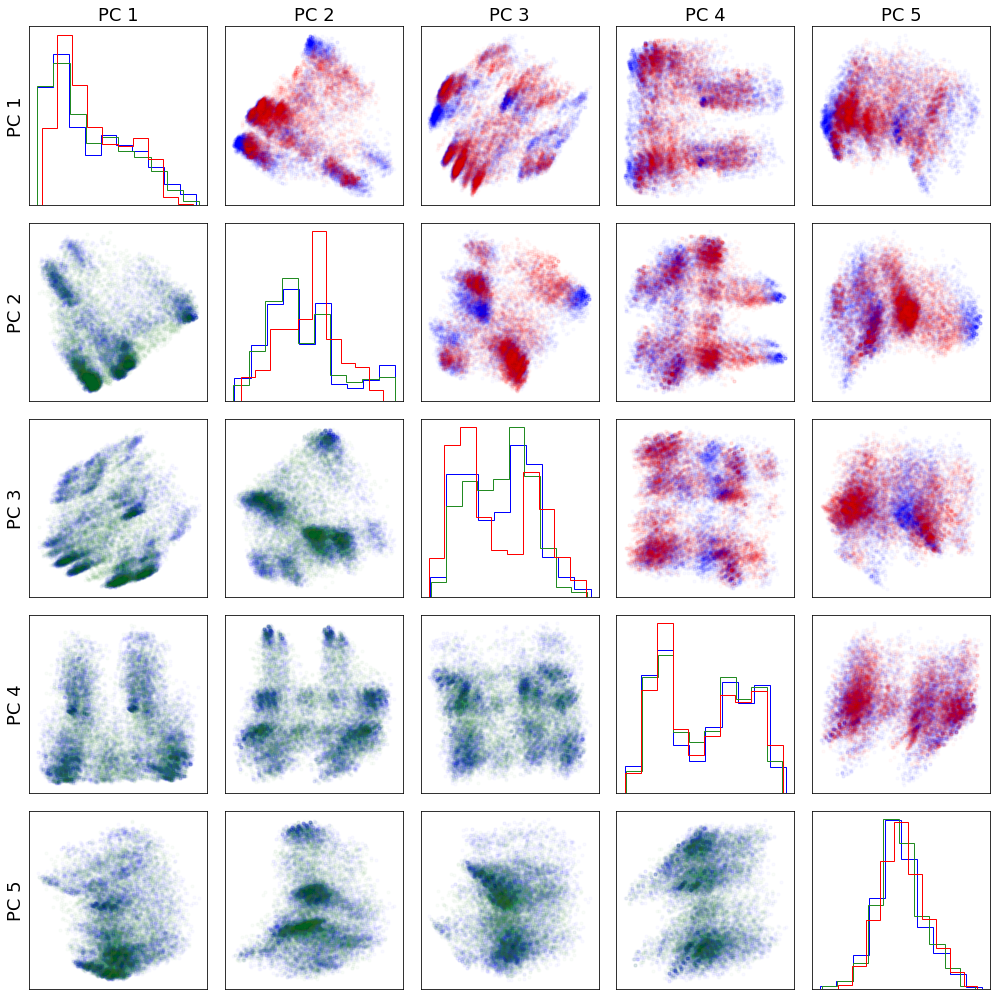}
\caption{Principal component analysis for the Socio case (numerical variables are converted to categorical). The data are projected to the first 5 principal components (PCs) of the training set (blue) and synthetic populations produced by VAE (green) and BN-exact (red). Diagonal plots show histograms of the data projected on the corresponding PC. On the off-diagonal plots, the PC projections are plotted one against each other, where each dot represents a data point projected on the two PCc.
}
\label{fig:pca2}
\end{figure}

\begin{figure}[ht!]
\centering
\includegraphics[width=0.9\linewidth]{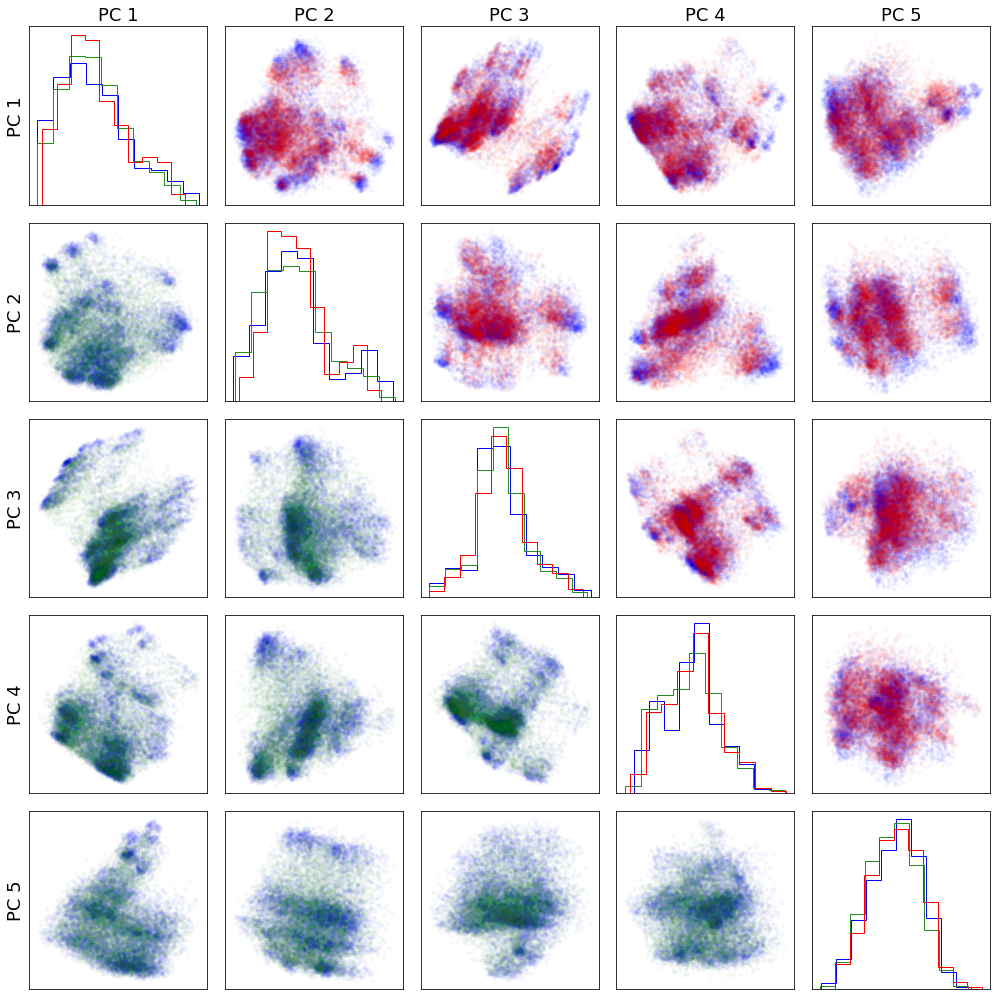}
\caption{Principal component analysis for the Extended case (numerical variables are converted to categorical). The data are projected to the first 5 principal components (PCs) of the training set (blue) and synthetic populations produced by VAE (green) and BN-tree (red). Diagonal plots show histograms of the data projected on the corresponding PC. On the off-diagonal plots, the PC projections are plotted one against each other, where each dot represents a data point projected on the two PCc.
}
\label{fig:pca3}
\end{figure}

For the Basic case, the data projections shown in Fig.~\ref{fig:pca1} suggest that all the three pools of agents have very similar clustering patterns. For the Socio case depicted in Fig.~\ref{fig:pca2}, the agents generated by the BN are in a worse agreement to the training data comparing to the agents generated by the VAE. It can be seen from both the histograms and the two-dimensional projections, especially for the first three PCs. For the Extended case (Fig.~\ref{fig:pca3}), the first two PCs of the data generated by the BN are slightly more off comparing to the VAE. However, it is more difficult to draw conclusions in this case using the visual inspection only. Here, it might be more appropriate to quantitatively compare the empirical distributions for different projections, for example, estimated as multidimensional histograms, similar to \ref{sec:aapendixB}, however, not considered in the current paper.

\section*{References}

\bibliographystyle{elsarticle-harv}
\bibliography{population_synthesis.bib}







\end{document}